\documentclass[journal]{IEEEtran}
\ifCLASSINFOpdf
\else
\fi
\usepackage{algorithm,setspace}
\usepackage{algorithmicx}
\usepackage{algpseudocode}
\usepackage{textcomp}
\usepackage{stfloats}
\usepackage{url}
\usepackage{verbatim}
\usepackage{graphicx}
\usepackage{graphicx}
\usepackage{multirow}
\usepackage{graphicx}
\usepackage{cite}
\usepackage{orcidlink}
\usepackage{hyperref}
\usepackage{amsfonts}       
\usepackage{nicefrac}       
\usepackage{microtype}      
\usepackage{lipsum}
\usepackage{amsmath}
\usepackage{fancyhdr}       
\usepackage{graphicx}       
\graphicspath{{media/}}     
\usepackage{xcolor}
\usepackage{subfigure}

\begin{document}
%
\title{Cyber Mobility Mirror: A Deep Learning-based Real-World Object Perception Platform Using Roadside LiDAR}

%

\author{
Zhengwei~Bai$^{\orcidlink{0000-0002-4867-021X}}$,~\IEEEmembership{Student Member,~IEEE,} 
Saswat~P.~Nayak, 
Xuanpeng~Zhao, 
Guoyuan~Wu$^{\orcidlink{0000-0001-6707-6366}}$,~\IEEEmembership{Senior Member,~IEEE,} 
Matthew~J.~Barth$^{\orcidlink{0000-0002-4735-5859}}$,~\IEEEmembership{Fellow,~IEEE,}
Xuewei Qi,~\IEEEmembership{Member, IEEE,}
Yongkang Liu, 
Emrah Akin Sisbot,
Kentaro Oguchi
\thanks{Zhengwei Bai, Saswat~P.~Nayak, Xuanpeng~Zhao, Guoyuan~Wu, and Matthew~J.~Barth are with the Department of Electrical and Computer Engineering, the University of California at Riverside, Riverside, CA 92507 USA (e-mail: zbai012@ucr.edu).}
\thanks{Xuewei Qi, Yongkang Liu, Emrah Akin Sisbot, and Kentaro Oguchi are with the Toyota North America R\&D Labs, Mountain View, CA 94043, USA.}}

\maketitle

\begin{abstract}
Object perception plays a fundamental role in Cooperative Driving Automation (CDA) which is regarded as a revolutionary promoter for the next-generation transportation systems. However, the vehicle-based perception may suffer from the limited sensing range and occlusion as well as low penetration rates in connectivity. In this paper, we propose \textit{Cyber Mobility Mirror} (\textit{CMM}), a next-generation real-time traffic surveillance system for 3D object perception and reconstruction, to explore the potential of roadside sensors for enabling CDA in the real world. The CMM system consists of six main components: 1) the data pre-processor to retrieve and preprocess the raw data; 2) the roadside 3D object detector to generate 3D detection results; 3) the multi-object tracker to identify detected objects; 4) the global locator to map positioning information from the LiDAR coordinate to geographic coordinate using coordinate transformation; 5) the cloud-based communicator to transmit perception information from roadside sensors to equipped vehicles, and 6) the onboard advisor to reconstruct and display the real-time traffic conditions via Graphical User Interface (GUI). In this study, a field-operational system is deployed at a real-world intersection, University Avenue and Iowa Avenue in Riverside, California to assess the feasibility and performance of our CMM system. Results from field tests demonstrate that our CMM prototype system can provide satisfactory perception performance with 96.99\% precision and 83.62\% recall. High-fidelity real-time traffic conditions (at the object level) can be geo-localized with an average error of $0.14m$ and displayed on the GUI of the equipped vehicle with a frequency of $3-4 Hz$.
\end{abstract}


%
\IEEEpeerreviewmaketitle

\section{Introduction}
With the rapid growth of travel demands, the transportation system is facing increasingly serious traffic-related challenges, such as improving traffic safety, mitigating traffic congestion, and reducing mobile source emissions. 
Taking advantage of recent strides in advanced sensing, wireless connectivity, and artificial intelligence, Cooperative Driving Automation (CDA) is attracting more and more attention over the past few years and is regarded as a transformative solution to the aforementioned challenges~\cite{fagnant2015preparing}. In the past few decades, several projects or programs have been conducted to explore the feasibility and potential of CDA. For instance, the California PATH program showed throughput improvement by a fully connected and automated platoon~\cite{misener2006path}. In the European DRIVE C2X project, the cooperative traffic system was assessed by large-scale field operational tests for various connected vehicle applications~\cite{stahlmann2011starting}. Recently, the U.S. Department of Transportation is leading the CARMA Program~\cite{carma} for research on CDA, leveraging emerging capabilities in both connectivity and automation to enable cooperative transportation system management and operations (TSMO) strategies. Additionally, the Autonet2030 Program led by EUCar is working on Cooperative Systems in Support of Networked Automated Driving by 2030~\cite{eucar_2021}.
\begin{figure}[!t]
    \centering
    \includegraphics[width=0.5\textwidth]{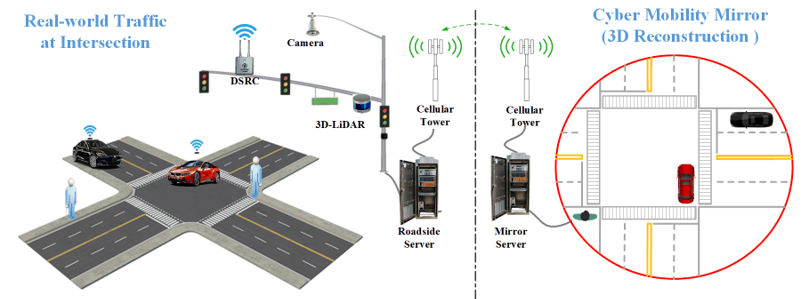}
    \caption{Illustration for CMM concept at an intersection scenario.}
    \label{fig:CMM concept}
\end{figure}
However, most of the aforementioned projects assume an ideal scenario, i.e., all vehicles are connected and automated. Because the presence of mixed traffic (with different types of connectivity and levels of automation) would be the norm, in the long run, one of the popular ways to enhance CAVs' adaptability in such a complicated environment is to improve their situation-awareness capability. For example, vehicles are equipped with more and more high-resolution onboard sensors and upgraded with powerful onboard computers to better perceive the surroundings and make decisions by themselves, a similar path to highly automated vehicles (HAVs)~\cite{arnold2019survey}. However, this roadmap is facing a couple of major challenges: 1) the cost of large-scale real-world implementation is prohibitive; and 2) the detection ranges are limited for onboard sensors, which also suffer from occlusion partially due to mounting heights and positions~\cite{yurtsever2020survey}.

Recently, roadside sensor-assisted perception is attracting a significant amount of attention for CAVs and is regarded as a promising way to unlock numerous opportunities for cooperative driving automation applications~\cite{bai2022infrastructure}. Current roadside sensing systems are mainly camera-based, which are cost-effective and well-developed for traffic surveillance such as turning movement counts, but hard to provide reliable object-level high-fidelity 3D information due to lighting conditions and shadow effects~\cite{datondji2016survey}.
    
Considering its capability to determine an accurate 3D location based on the point cloud data, LiDAR gets more popular in infrastructure-based traffic surveillance. Previous studies validated the performance of roadside LiDAR for vehicle detection~\cite{wu2020automatic}, vehicle tracking~\cite{zhao2019detection}, lane identification~\cite{wu2018automatic}, pedestrian near-crash warning~\cite{wu2020improved}, and other applications~\cite{zhang2019automatic,lv2019lidar}. These studies laid the foundation for applications with roadside LiDAR-based perception systems. However, all these systems deploy a traditional perception pipeline ~\cite{song2020background, ester1996density}, consisting of background filtering, point cloud clustering, object classification, and object tracking. Such pipeline may generate stable results but suffer from uncertainties and generality~\cite{zou2019object}. With the development of computer vision, deep learning-based perception models show great potential to overcome the above issues. However, very few studies applied deep learning-based perception algorithms to roadside LiDAR systems.

The main contributions of this paper can be summarized as follows: 
\begin{enumerate}
    \item To the best of the authors' knowledge, this paper is the first attempt to comprehensively build a deep learning-based real-world platform, called \textit{Cyber Mobility Mirror (CMM)}, for 3D object-level detection and tracking at a signalized intersection using the roadside LiDAR. 
    \item To improve the transferability of learning-based detection models, Roadside Point-cloud Encoder and Decoder (RPEaD) is proposed.
    \item A real-time 3D multi-object tracking method, called \textit{3DSORT}, is developed.
    \item A geo-localization method is proposed to support the reconstruction from the detector output to geographic visualization.
\end{enumerate}
The CMM platform can serve as the stepping stone to enabling various cooperative driving automation (CDA) applications.

The rest of this paper is organized as follows: related work is firstly introduced in Section~\ref{background}. Section~\ref{CMM} shows the concept and structure of CMM, followed by a detailed description of the associated field operational system in Section~\ref{FOS}. The results and analyses are discussed in Section~\ref{cmm results} and the last section concludes this paper with further discussion.

\section{Background}
\label{background}
Situation awareness is one of the fundamental building blocks for Driving Automation (DA). Specifically, 3D object detection and tracking play a crucial role in perceiving the environment. Meanwhile, traffic object reconstruction helps drivers better understand the traffic conditions. Hence, in this section, related work about detection, tracking, and reconstruction of traffic objects is presented, based on a detailed literature review. 
\subsection{Traffic Object Detection}
Object detection is a fundamental task of environment perception and has also gone through a rapid development process in the past several decades. Back twenty years ago, a vision-based traffic detection system made an impressive achievement using statistical methods~\cite{cucchiara2000statistic}. For instance, Aslani and Mahdavi-Nasab~\cite{aslani2013optical} proposed an optical flow-based moving object detection method for traffic surveillance. However, these model-based methods cannot provide high fidelity detection results for more delicate applications, e.g., precise localization and object-level tracking. To explore highly accurate moving object detection methods, researchers started applying artificial neural networks~\cite{huang2013highly}. 

With the tremendous progress of convolutional neural networks (CNNs) in vision-based tasks, CNN-based object detection methods have attracted a significant amount of attention in traffic surveillance~\cite{boukerche2021object}. For instance, \textit{You Only Look Once} (\textit{YOLO})~\cite{redmon2016you} and its variants, due to an impressive performance in real-time multi-object detection, get very popular in high-resolution traffic monitoring scenarios. For multi-scale vehicle object detection, Mao et al.~\cite{mao2020finding} added \textit{Spatial Pyramid Pooling} (\textit{SPP}) modules in YOLO to obtain multi-resolution information. The \textit{Single Shot MultiBox Detector} (SSD)~\cite{liu2016ssd} is also of significance in traffic applications. Based on SSD, Wang et al.~\cite{wang2018multi} proposed a novel multi-object detection model to improve the overall perception performance under a variety of traffic scenarios based on a multi-kernel CNN. \textit{Faster RCNN}~\cite{ren2015faster} is another generic epoch-making detection method, utilizing the region proposal ideology. To further improve the object detection performance for Faster-RCNN, Li et al.~\cite{li2021method} proposed a cross-layer fusion structure based on Faster RCNN to achieve a nearly 10\% higher average accuracy in complex traffic environments, e.g., dense traffic with shadows and occlusions. 

Except for the general object detection task applied in traffic scenes, many studies focus on specific perception cases. For instance, considering that existing traffic surveillance systems were made up of costly equipment with complicated operational procedures, Mhalla et al.~\cite{mhalla2018embedded} designed an embedded computer-vision system for multi-object detection in traffic surveillance. For small object detection, Lian et al.~\cite{lian2021small} proposed an attention feature fusion block to better integrate contextual information from different layers that could achieve much better performance. Targeting other edge situations, i.e.,  highly crowded traffic scenarios, Gahlert et al.~\cite{gahlert2020visibility} proposed the \textit{Visibility Guided Non-Maximum Suppression} (vg-NMS) to improve the detection accuracy by leveraging both pixel-based object detection and \textit{amodal perception} paradigms. For situation awareness, Guindel et al.~\cite{guindel2018fast} proposed a deep CNN to jointly handle object detection and viewpoint estimation.

To support object-level cooperative operations, detecting the objects in a 3D format is a straightforward and promising way for high-fidelity situation awareness. Hence, owing to the capability of generating 3D point clouds with spatial information, it is increasingly popular for deploying 3D LiDAR to traffic environment perception. Wu et al. \cite{wu2020automatic}, proposed a revised \textit{Density-Based Spatial Clustering of Applications with Noise} (3D-DBSCAN) method  to detect vehicles based on roadside LiDAR sensors under rainy and snowy conditions. Using a roadside LiDAR, Zhang et al. proposed a three-stage inference pipeline, called \textit{GC-net}~\cite{zhang2020gc}, including the gridding, clustering, and classification. In this study, the raw point cloud data (PCD) was firstly mapped into a grid structure and then clustered by the \textit{Grid-Density-Based Spatial Clustering} algorithm. Finally, a CNN-based classifier was applied to categorize the detected objects by extracting their local features. Liu et al. proposed a roadside LiDAR-based object detection approach by following the conventional background filtering and clustering pipeline~\cite{Liu9434525}, where point correlation with \textit{KDTree}~\cite{redmond2007method} neighborhood searching and adaptive Euclidean clustering was applied, respectively. To distinguish the moving object from the point cloud, Song et al. proposed a hierarchical searching method based on the feature distribution of point clouds to achieve background filtering and object detection~\cite{Song9216093}. Although 3D LiDAR has innate advantages to deal with 3D object detection, the lack of labeled roadside dataset significantly limits the potential for applying deep leaning-based detectors to roadside LiDAR sensors. Hence in this paper, an point cloud encoder-decoder method is proposed to enable the detection model to work on roadside point cloud with training on onboard dataset.

\subsection{Traffic Object Tracking}
Deploying CDA in urban environments poses a series of difficult technological challenges, out of which object tracking is arguably one of the most significant since it provides the identification information for other subsequent technical models~\cite{2021SAE}. Object tracking can be classified into two categories in terms of the number of objects tracked at one time: one is single-object tracking (SOT) and the other is multi-object tracking (MOT). SOT has been investigated over several decades and the \textit{Kalman filtering} methods or \textit{particle filtering} methods have been employed widely~\cite{cuevas2005kalman, okuma2004boosted} for this type of task. For MOT tasks, some approaches have been proposed with the focus on improving accuracy and real-time performance. For instance, Bewley et al.~\cite{bewley2016simple} proposed \textit{Simple Online and Real-time Tracking} (SORT) that can achieve MOT in a high frame rate without much compromising the accuracy. Based on the structure of SORT, Nicolai et al.~\cite{8296962} proposed a multi-object tracker -- \textit{DeepSORT}, which was capable of tracking objects with longer periods of occlusions and effectively reducing the number of identity switches by integrating the appearance features. However, Deep SORT does not apply to 3D objects. 

Considering the evolution of sensor technology and perception methods, camera-based approaches play a dominant role in traffic object tracking over the past several decades. For instance, Aslani et al.\cite{aslani2013optical} applied the \textit{Optical Flow} algorithm to detect and track moving objects by the intensity changes of frames. To improve the MOT performance, Fernandez-Sanjurjo et al.~\cite{fernandez2019real} built a real-time traffic monitoring system and data association with the \textit{Hungarian} algorithm. Based on cameras equipped on an unmanned aerial vehicle (UAV), researchers~\cite{zhao2019detection, balamuralidhar2021multeye} applied correlation filters\cite{bolme2010visual} to MOT tasks. 
Chen et al. proposed a camera-based edge traffic flow monitoring scheme using DeepSORT~\cite{chen2020edge}. Recent advances in LiDAR technology enable it to hold a place in traffic object tracking tasks, by leveraging the point cloud data. For instance, Cui et al.~\cite{cui2019automatic} provided a simple \textit{global nearest neighbor} (GNN) method to track multiple vehicles based on the spatial distance between consecutive frames. \textit{Adaptive probabilistic filtering} was utilized by Kampker et al.~\cite{kampker2018towards} to handle uncertainties due to sensing limitations of 3D LiDARs and the complexity of targets' movements. Zhang et al.,~\cite{zhang2020vehicle} used \textit{unscented Kalman filter} (UKF) and joint probability data association filter for MOT, which improved the accuracy of estimated vehicle speed through an image matching process.

\subsection{Traffic Object Reconstruction}

Traffic reconstruction, traditionally, means rebuilding the traffic scenarios or parameters based on recorded sensor data, such as loop detectors and surveillance cameras ~\cite{herrera2007traffic, jiang2018compressive}. These traffic-level reconstruction data are valuable for macroscopic traffic management. In this paper, nevertheless, the object-level reconstruction means rebuilding the 3D location or shape of certain objects based on sensor data, which can more concrete information to support subsequent CDA applications. Several studies have been conducted in this emerging area. Cao et al.~\cite{9107468} developed a camera-based 3D object reconstruction method on the Internet of Vehicles (IoV) environment. Rao and Chakraborty~\cite{9151364} proposed a LiDAR-based monocular 3D shaping to reconstruct the surrounding objects for onboard display, which has a similar purpose to the reconstruction work in this paper.


\section{Cyber Mobility Mirror (CMM)}
\label{CMM}
To explore the potential of the roadside sensing system, we propose a novel infrastructure-based object-level perception system, named \textit{Cyber Mobility Mirror}. In this section, the core concept of CMM and the associated platform implemented in the real world are introduced.

\begin{figure}[!ht]
    \centering
    \includegraphics[width=0.5\textwidth]{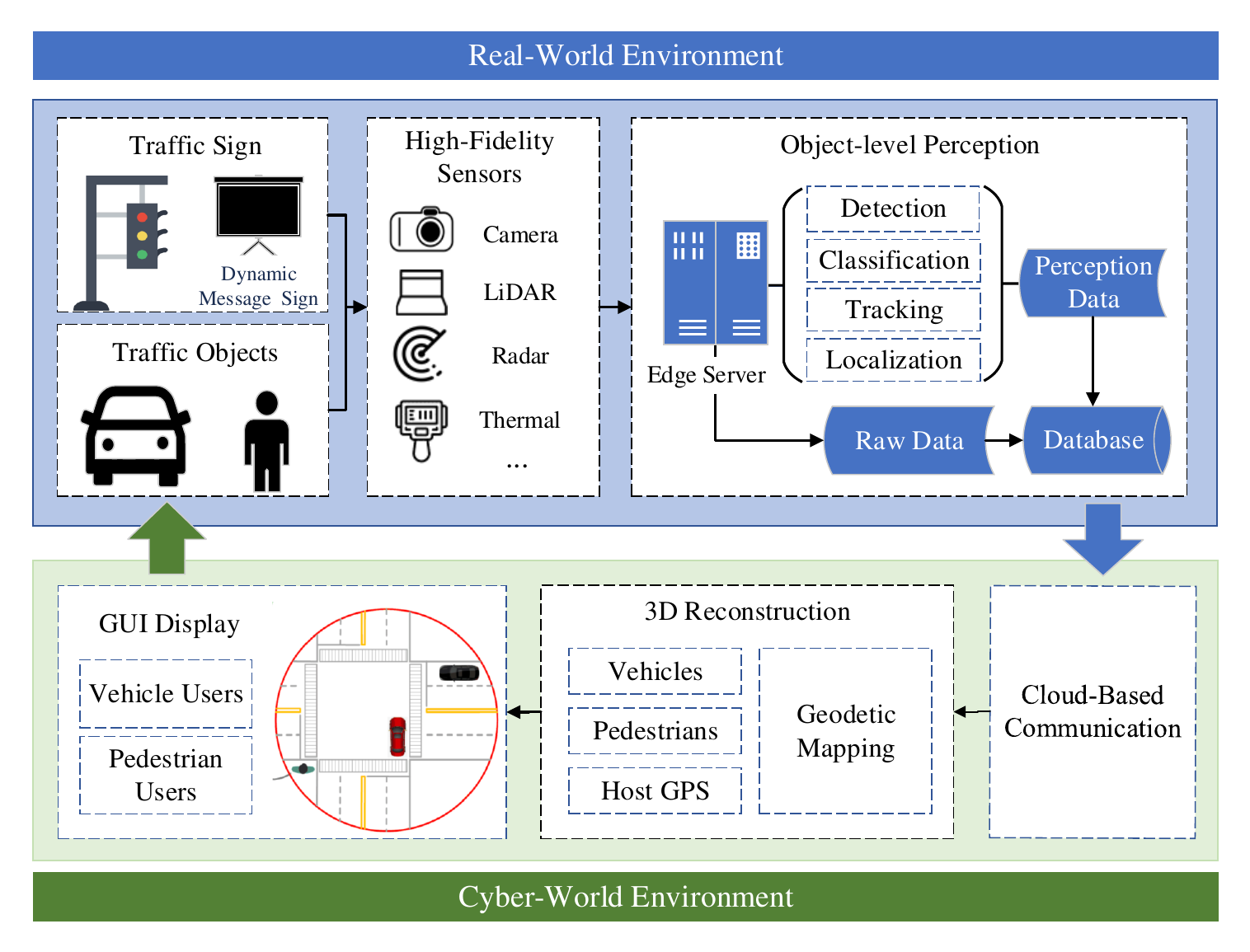}
    \caption{Systematic diagram for the core concept of CMM.}
    \label{fig:cmm core concept}
\end{figure}

\subsection{Core Concept of CMM}
CMM aims to enable real-time object-level traffic perception and reconstruction to empower various cooperative driving automation (CDA) applications, such as \textit{Collision Warning}~\cite{wu2020improved}, \textit{Eco-Approach, and Departure} (EAD)~\cite{bai2022hybrid}, and \textit{Cooperative Adaptive Cruise Control} (CACC)~\cite{wangCACC}. In the CMM system, traffic conditions (i.e., ``\textit{mobility}'') are detected by high-fidelity sensors and advanced perception methods, such as object detection, classification, and tracking. In the ``\textit{cyber}'' world, digital replicas (i.e., ``\textit{mirrored}'' objects) are built to reconstruct the traffic in \textit{real-time} via high-definition 3D perception information, such as the detected objects' geodetic locations (rendered on the satellite map), 3D dimensions, speeds, and moving directions (or headings). Then, this ``\textit{mirror}'' can act as the perception foundation for numerous CDA applications in a real-world transportation system. 

Specifically, Fig.~\ref{fig:cmm core concept} illustrates the system diagram for the core concept of CMM. Traffic objects can be detected by high-fidelity sensors equipped on the infrastructure side and the sensing data is processed by an edge server to generate object-level information and enable various functions, such as detection, classification, tracking, and geodetic localization. The perception information is also transmitted to a cloud server for distribution and 3D reconstruction. The reconstructed traffic environment can be displayed on the GUI of connected road users to support various CDA applications. 

\begin{figure*}[!hb]
    \centering
    \includegraphics[width=0.8\textwidth]{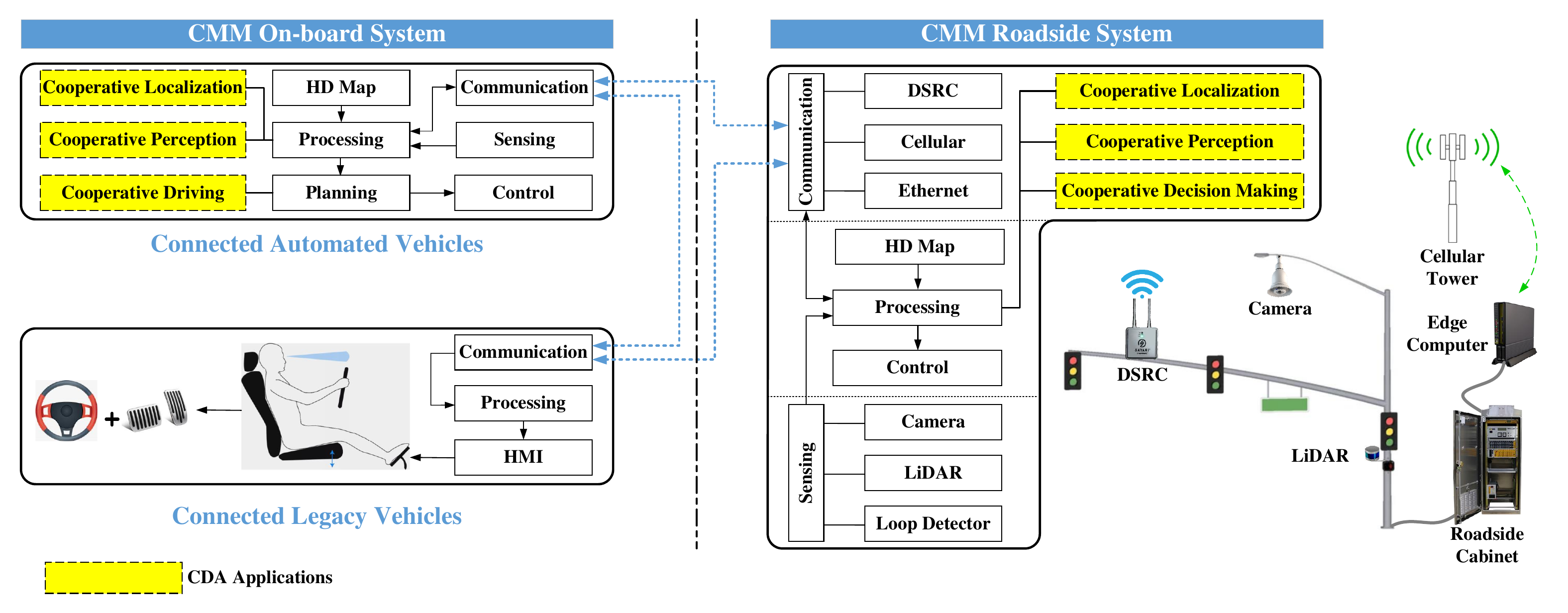}
    \caption{System structure for CMM system in Real-World Traffic Environment.}
    \label{fig:cmm realworld}
\end{figure*}

\subsection{Systematic Structure of CMM}
In the real-world traffic environment, the system architecture of the CMM system is designed by following the core concept. Specifically, the CMM system can be divided into two main parts: the CMM Roadside System (CRS) and the CMM Onboard System (COS). Fig.~\ref{fig:cmm realworld} illustrates the system architecture. CRS and COS are introduced in detail as follows:

\subsubsection{CMM Roadside System}
CRS consists of 1) roadside sensors, e.g., LiDAR in this study, to perceive traffic conditions and generate high-fidelity sensor data; 2) edge computing-based real-time perception pipeline to achieve sensor fusion (if appropriate), object detection, classification, and tracking tasks; and 3) communication devices to receive information from other road users, infrastructure or even "clouds", and share perceived results with them via different kinds of protocols.

\subsubsection{CMM Onboard System}
For CAVs, COS can receive the object-level perception data from CRS and then act as the perception inputs to support various CDA applications, such as CACC, cooperative merging, cooperative eco-driving; and for Connected Human-driven Vehicles (CHVs), COS can also provide them with real-time traffic information via the human-machine interface (HMI) to improve driving performance or to avoid possible crashes due to occlusion.

In this paper, the CMM concept is implemented in the real world and a field operational system is developed for real-world testing, which will be discussed in Section~\ref{FOS}.

\section{CMM Field Operational System}
\label{FOS}

\subsection{System Overview}
\begin{figure*}[!ht]
    \centering
    \includegraphics[width=0.8\textwidth]{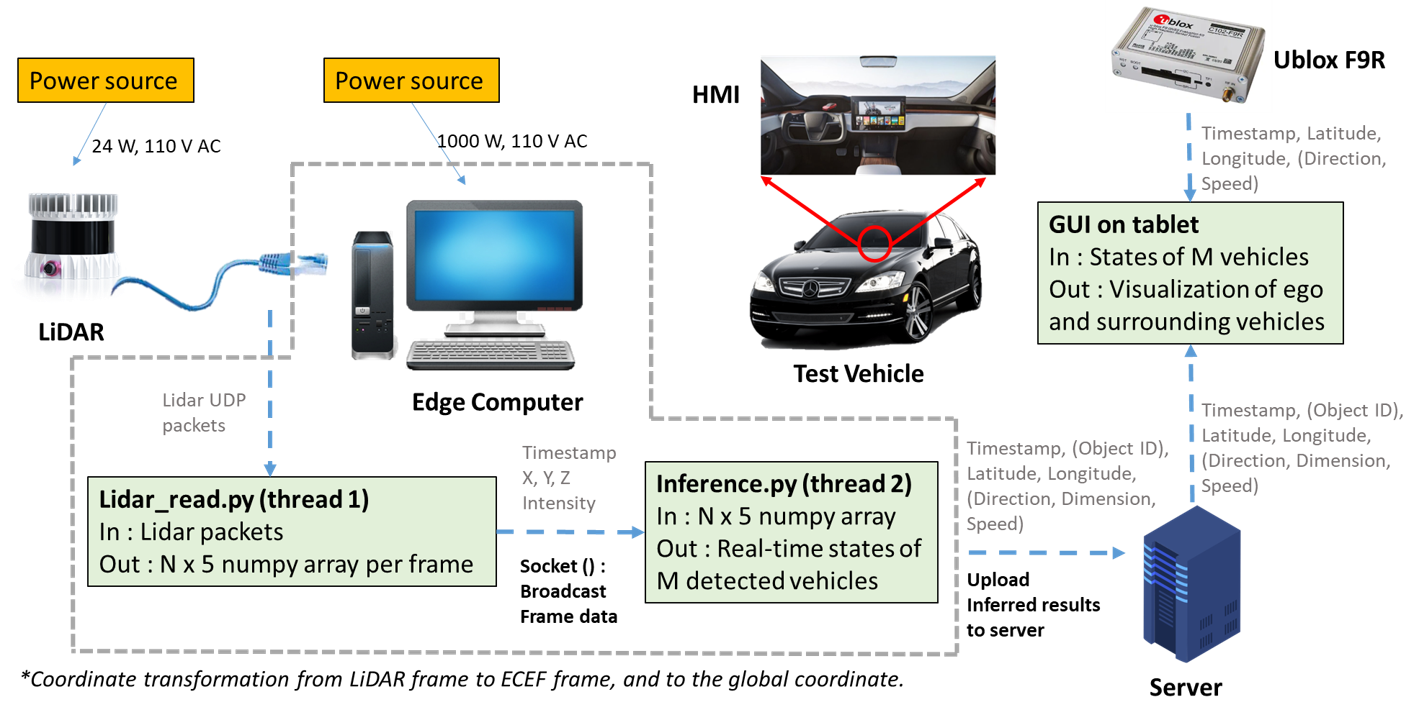}
    \caption{The architecture for CMM field operational prototype system.}
    \label{fig:CMM_FOS}
\end{figure*}

The system overview for the CMM Field Operational System (FOS) is shown in Fig.~\ref{fig:CMM_FOS}. The FOS mainly consists of a roadside 3D LiDAR for data collection, an edge-computing system for data processing, a cloud server for data distribution, and a test vehicle equipped with connectivity and Graphic User Interface (GUI). To be specific, the LiDAR is installed on the signal pole high enough to achieve better coverage. The edge computer retrieves 3D point cloud data from the roadside LiDAR and then generates high-definition perception information (i.e., 3D object detection, classification, and tracking results) which is transmitted to the cloud server via Cellular Network. A CHV equipped with the CMM OBUs (including a GPS receiver, onboard communication device, and a tablet) can receive the perception information, and reconstruct and display the object-level traffic condition on GUI in real-time.

\subsection{System Initialization}

\begin{figure}[!h]
    \centering
    \includegraphics[width=0.5\textwidth]{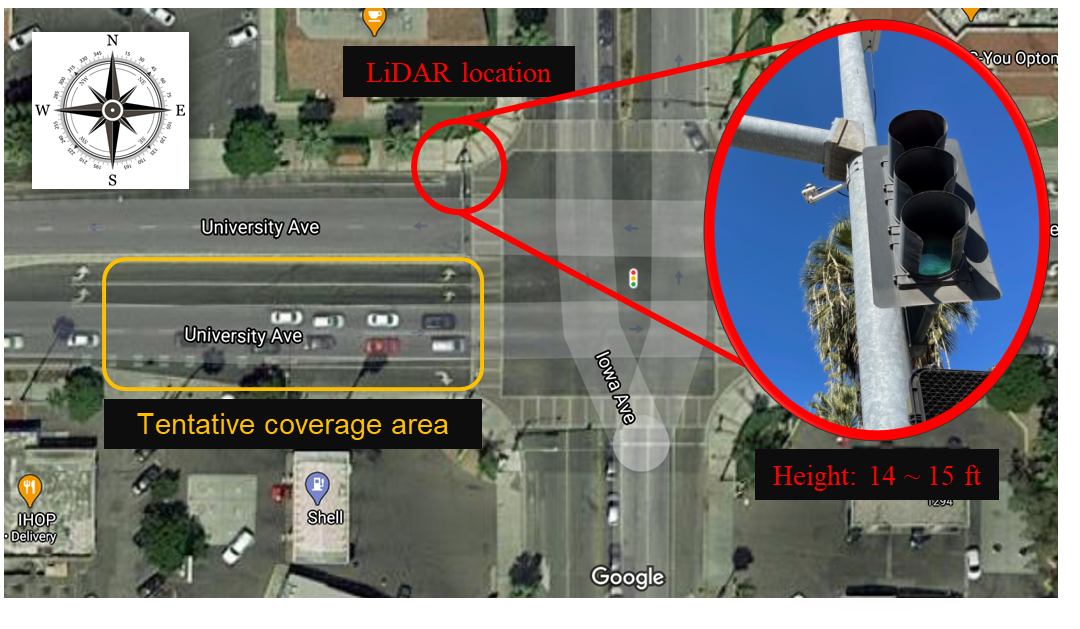}
    \caption{Location and installation of the equipped roadside LiDAR.}
    \label{fig:LiDAR install}
\end{figure}

As demonstrated by Fig.~\ref{fig:LiDAR install}, the LiDAR is installed at the northwest corner of the intersection (marked as the red circle) of University Ave. and Iowa Ave. in Riverside, California. In this work, an OUSTER\textregistered 64-Channel 3D LiDAR is used as a major roadside sensor, mounted on a signal pole at the height of 14-15 ft above the ground with the appropriate pitch and yaw angles to cover the monitoring area enclosed by the orange rectangle in Fig.~\ref{fig:LiDAR install}. The edge computer at the intersection receives the stream of LiDAR data in the form of UDP packets. Other point cloud attributes such as 3D location, i.e., $x$, $y$, $z$, and the intensity, $i$, of each point are bundled into an $N\times4$ array to be used in the inference pipeline.

\subsection{Data Retrieving and Preprocessing}

The raw point cloud data is generated by a 64-channel 3D LiDAR and then the edge computer retrieves the raw data through an Ethernet cable via UDP communication. In this paper, the detection range $\Omega$ for the roadside LiDAR is defined as a $102.4m \times 102.4m$ area centered on the location of LiDAR. The raw point cloud data can be described by:
\begin{equation}
    \mathcal{P} = \{[x, y, z, i]\, |\, [x, y, z] \in \mathbb{R}^{3}, i\in [0.0, 1.0] \}.
\end{equation}
Then, $\mathcal{P}$ is geo-fenced by:
\begin{equation}
\mathcal{P}_{\Omega} = \{[x, y, z, i]^{T} \, |\, x \in \mathcal{X}, y \in \mathcal{Y}, z \in \mathcal{Z}\}   
\end{equation}
where $\mathcal{P}_{\Omega}$ represents the 3D point cloud data after geofencing; and $\mathcal{X}$ and $\mathcal{Y}$ are set as $[-51.2m, 51.2m]$. Considering the calibrated height of the roadside Lidar to be $4.74m$, $\mathcal{Z}$ is set as $[-5.0m, 0m]$. 

\subsection{3D Object Detection from roadside LiDAR}
\subsubsection{Roadside Point-cloud Encoder and Decoder}

\begin{figure}[!h]
    \centering
    \includegraphics[width=0.5\textwidth]{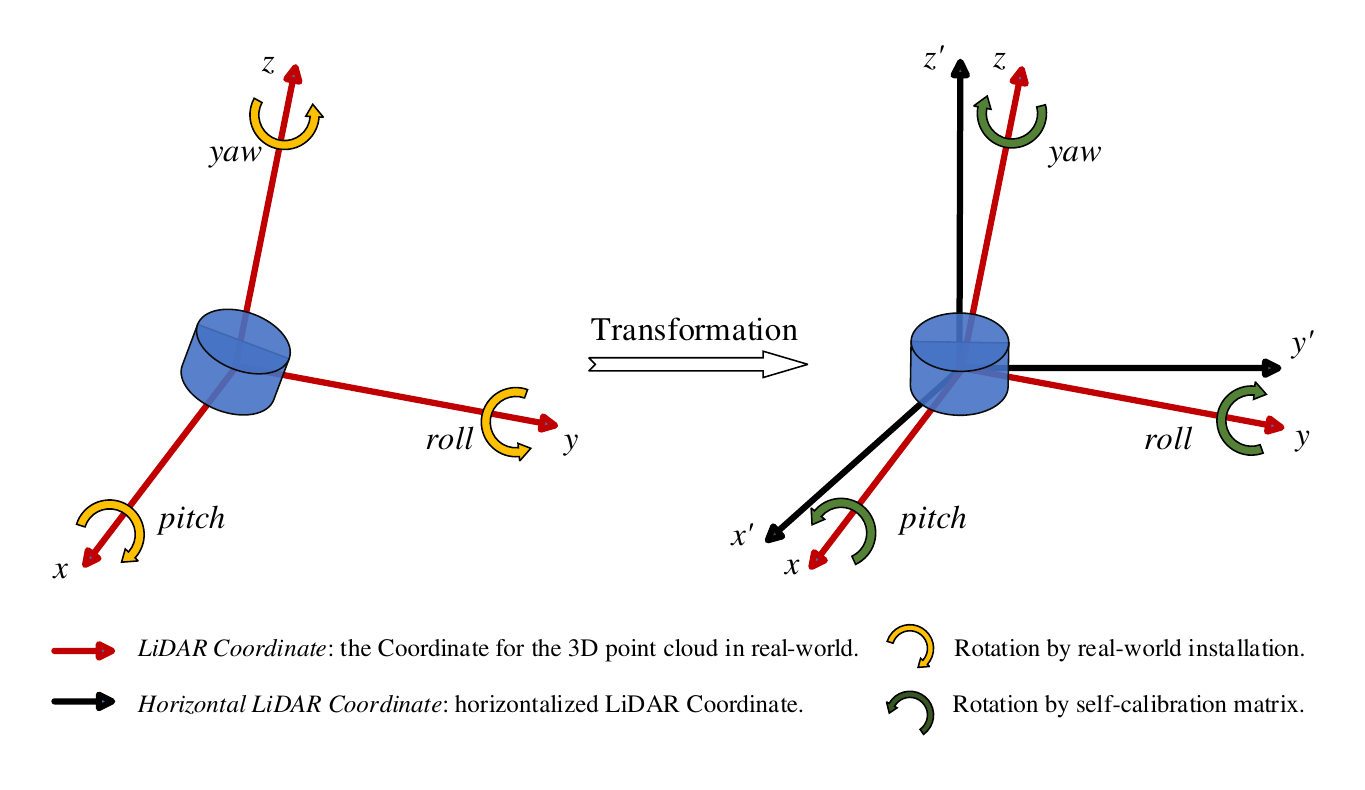}
    \caption{Description of the initial transformation for LiDAR point cloud data.}
    \label{fig:carla-kitti-coor}
\end{figure}

Considering the LiDAR's limited vertical field of view (FOV), it is installed with an adjusted rotation angle including pitch, yaw, and roll to cover the desired surveillance area as shown in Fig~\ref{fig:carla-kitti-coor}. To build the system cost-effectively, we try to use an open-source dataset to train our detection model, e.g., Nuscenes~\cite{caesar2020nuscenes}. However, these available datasets are collected based on a vehicle-equipped LiDAR. These LiDAR sensors have different spatial configurations from ours and the model trained on these datasets may not work well for our roadside point clouds. 

To empower the model with the capability of training on onboard datasets while inferencing on the roadside, we propose the Roadside Point-cloud Encoder and Decoder (RPEaD). The main purpose of RPEaD is to transform roadside point clouds into a space in which the model trained on the onboard datasets can work out. The transformation process of the encoder is described in Fig.~\ref{fig:carla-kitti-coor}.

To achieve the transformation, we propose a self-calibration approach for the roadside-LiDAR pose by using Least Square Regression (LSR) to the point clouds. The coordinate for roadside point clouds are defined as \textit{LiDAR Coordinate (L-Coor)} and the coordinate of point clouds after encoding, is defined as \textit{Horizontal Coordinate (H-Coor)}. Using LSR, the least square plane is generated to represent the $x-y$ plane of the \textit{L-Coor}. Then the 3D rotation matrix can be generated as $\mathcal{P}_{Cali}$, which is shown as:
\begin{equation}
\mathcal{P}_{Cali} = \begin{bmatrix}
a &  b&  c\\
d &  e&  f\\
g &  h&  i
\end{bmatrix}
\end{equation}
where $a, ..., i$ are the parameters generated from LSR. For translation, the vertical offset $\Delta z$ is defined as:
\begin{equation}
    \Delta z = z_{roadside} - z_{onboard}
\end{equation}
where $z_{roadside}$ and $z_{onboard}$ represent the heights of the roadside LiDAR and the onboard LiDAR (used in the training dataset), respectively.

The whole encoding process is defined by:
\begin{equation}
\label{LcoorToHcoor}
    \mathcal{P}_{\mathcal{H}} = \mathcal{P}_{\Omega} \cdot \begin{bmatrix}
\mathcal{P}_{Cali}& 0\\
0&1
\end{bmatrix} + [0, 0, \Delta z, 0]
\end{equation}

\subsubsection{Object Detection Network}
\begin{figure}[!h]
    \centering
    \includegraphics[width=0.5\textwidth]{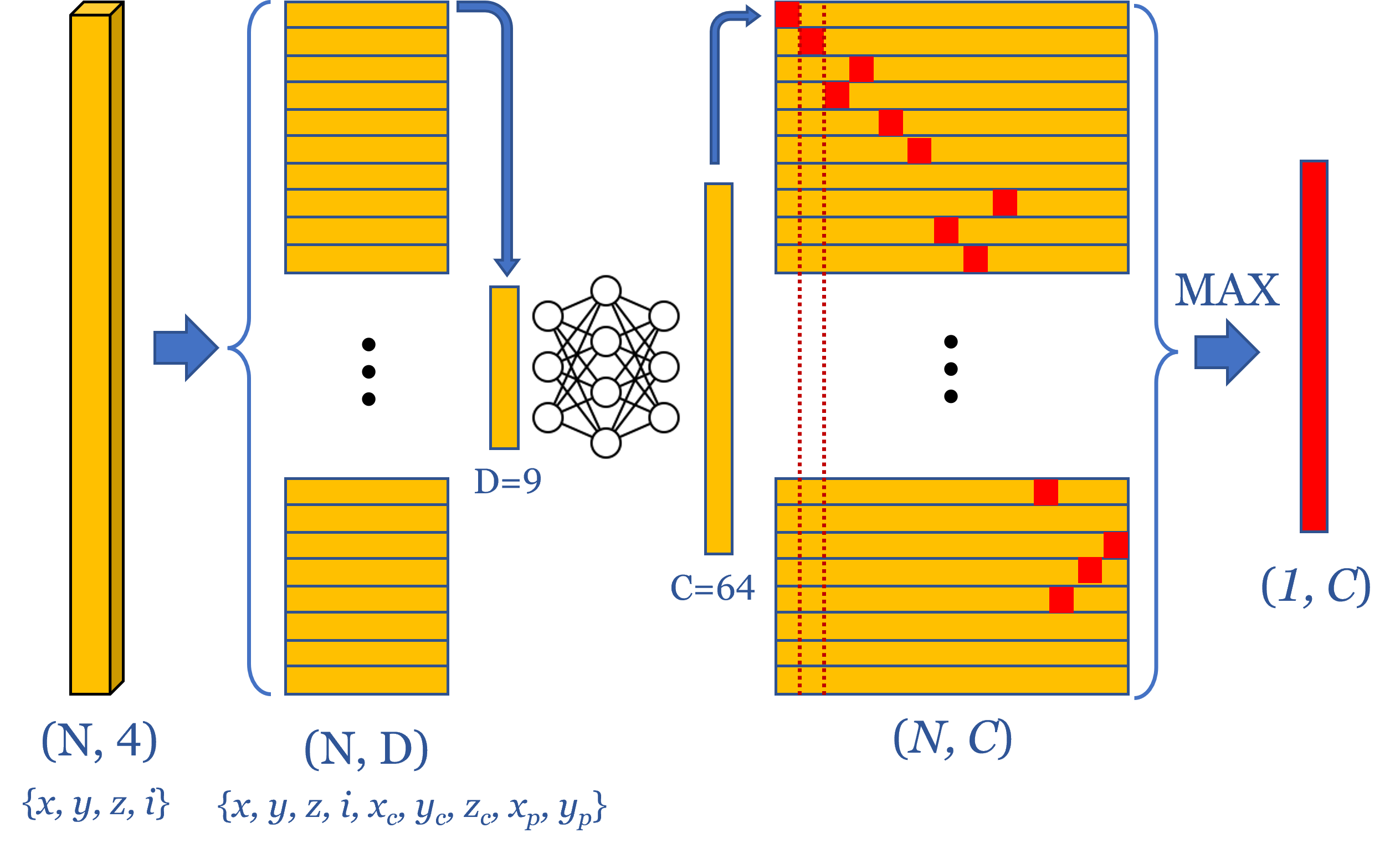}
    \caption{Process for the feature extraction and compression.}
    \label{fig:pp}
\end{figure}

Although the roadside point cloud is transformed into the coordinate suitable for training on the onboard dataset. The detection model has still required a large tolerance for the difference in data. Since there is a large shifting, i.e., near $3m$, along $z-$axis, to make the model not too sensitive for $z-$axis data, we voxelized the point cloud following the strategy applied in \cite{lang2019pointpillars}, i.e., only voxelization on the $x-y$ plane to generate point cloud pillars. Then data aggregation, as shown in Fig.~\ref{fig:pp}, is designed to extract and compress the features which will be sent to the deep neural network for generating predicted bounding boxes.

\begin{figure}[!h]
    \centering
    \includegraphics[width=0.5\textwidth]{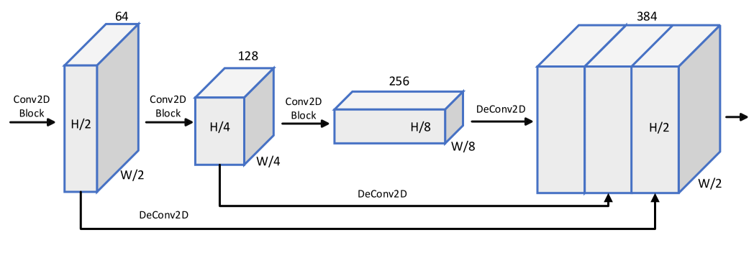}
    \caption{Deep neural network backbone for hidden feature extraction.}
    \label{fig:bbk}
\end{figure}

After the data aggregation, Fig~\ref{fig:bbk} shows the designed feature pyramid network (FPN) followed by a 3D anchor-based detection head~\cite{liu2016ssd} to generate predicted bounding boxes. 

For the loss functions, localization and classification are considered. To be specific, ground target (GT) and anchors are defined by a 8-dimensional vector $(x,y,z,w,l,h,\theta)$. The localization regression residuals between ground truth and anchors are defined by:

\begin{equation}
    \Delta x = \frac{x^{gt} - x^a}{d^a}, 
    \Delta y = \frac{y^{gt} - y^a}{d^a}, 
    \Delta z = \frac{z^{gt} - z^a}{h^a},
\end{equation}
\begin{equation}  
    \Delta w = \log\frac{w^{gt}}{w^a}, 
    \Delta l = \log\frac{l^{gt}}{l^a}, 
    \Delta h = \log\frac{h^{gt}}{h^a}, 
\end{equation}
\begin{equation}  
    \Delta \theta = sin(\theta^{gt} - \theta^a)
\end{equation}
where the superscript $gt$ and $a$ represent the ground truth and anchor, respectively, and $d^a$ is defined by: 
\begin{equation}
    d^a = \sqrt{(w^a)^2 + (l^a)^2}.
\end{equation}
The total localization loss is:
\begin{equation}
    \mathcal{L}_{loc} = \sum_{b \in (x, y, z, w, l, h, \theta)} \text{SmoothL1}(\Delta b)
\end{equation}
Inspired by~\cite{yan2018second}, a softmax classification loss, $\mathcal{L}_{dir}$, is used to distinguish flipped boxes.
The object classification is enabled by the focal loss \cite{lin2017focal}, which is shown as:
\begin{equation}
    \mathcal{L}_{cls} = -\alpha_{a}(1-p^a)^\gamma \log p^a,
\end{equation}
where $p^a$ is the class probability of an anchor, and $\alpha$ and $\beta$ are set as the same as the original paper. Hence, the total loss is:
\begin{equation}
    \mathcal{L} = \frac{1}{N_{pos}}(\beta_{loc}\mathcal{L}_{loc} + \beta_{cls}\mathcal{L}_{cls} + \beta_{dir}\mathcal{L}_{dir}),
\end{equation}
where $N_{pos}$ is the number of positive anchors and $\beta_{loc}$, $\beta_{cls}$ and $\beta_{dir}$ are set as $2$, $1$, and $0.2$.

\subsection{3D Multi-Object Tracking}
For real-time 3D MOT, we propose \textit{3DSORT} by adding 3D object matching on DeepSORT~\cite{chen2020edge}. To be specific, 2D location information is filtered from the 3D detection results, and the 2D location data is fed into the DeepSORT model to generate the 2D MOT results, i.e., unique identification (ID) number for each object. Then, a Euclidean distance-based 3D object matching algorithm is designed to generate the enhanced 3D MOT results. Algorithm~\ref{alg:deep sort} demonstrates the details of 3DSORT.

\begin{algorithm}
\setstretch{1.1}
\caption{The description for 3DSORT.}
\label{alg:deep sort}
\begin{algorithmic}[1] 
\Require The instant 3D object detection results: $Dobj = \{D^{(i)}(x, y, z, w, l, h, \theta)| i = 1, 2, ..., N_{Dbbx}\}$; 
\Ensure The multi-object tracking results: $Tobj = \{T^{(i)}(x, y, z, w, l, h, \theta, id)| i = 1, 2, ..., N_{Dbbx}\}$;
\Function{3D DeepSORT}{$Dobj$}
    \State $Dobj_{2d} \gets {D^{(i)}(x, y, w, l)| i = 1, 2, ..., N_{Dbbx}}$; 
    \State $Tobj_{2d} = \{T_{2d}^{(j)}(x, y, w, l, id)| j = 1, 2, ..., N_{Tbbx}\} \gets DeepSORT(Dobj_{2d})$;
    \For{$Dobj_{2d}^{(i)} \in Dobj_{2d}$}
        \For{$Tobj_{2d}^{(j)} \in Tobj_{2d}$}
            \If{Euclidean distance of 
            $(Dobj_{2d}^{(i)}, Tobj_{2d}^{(j)}) < d_o$}
            \State $T^{i} \gets [D^{(i)}, Tobj_{2d}(id)] $; Continue;
            \EndIf
        \EndFor
    \EndFor
    \State $Tobj = \{T^{(i)} | i = 1, 2, ..., N_{Dbbx}\}$
    \State \texttt{return} $Tobj$;
\EndFunction
\end{algorithmic}
\end{algorithm}
where $N_{Dbbx}$ and $N_{Tbbx}$ are the numbers of the detection bounding boxes and 2D tracking boxes, respectively. Additionally, $id$ represents the tracking identification number for each unique object. $d_o$ is the matching distance which is defined as $0.2m$.

\subsection{Geo-localization}
To endow the perception data with more generality, the geo-referencing of the point cloud is developed in this work. However, the output $T_{boxes}$ from the 3D MOT is calculated based on the Horizontal-LiDAR Coordinate, i.e., a Cartesian Coordinate centered with the sensor installed evenly. Thus, the input of the geo-localization data, i.e., the $T_{boxes}$ from Algorithm~\ref{alg:deep sort}, is then fed into a multi-step transformation process to transform the object location information to Geodetic Coordinate, i.e., latitude, longitude, and altitude. 
There are three steps: 1) from the horizontal-LiDAR coordinate to the real LiDAR coordinate; 2) from the real LiDAR coordinate to the Geocentric Earth-centered Earth-fixed (ECEF) coordinate; and 3) from ECEF coordinate to the geodetic coordinate (i.e., latitude, longitude, and altitude). Specifically, the World Geodetic System 1984 (WGS84) is applied for the geo-transformation. The transformation from the Horizontal LiDAR coordinate to the ECEF coordinate system is shown in Eq.~\ref{eq:geo-loc}. 
\begin{equation}
\label{eq:geo-loc}
\begin{bmatrix}
X_{ecef}\\ 
Y_{ecef}\\ 
Z_{ecef}\\
1
\end{bmatrix}^{T}
= \begin{bmatrix}
X_{hor}\\ 
Y_{hor}\\ 
Z_{hor}\\ 
1\end{bmatrix}^{T}
\cdot \mathcal{P}_{Cali}^{-1}\cdot \mathcal{P}_{ECEF} 
\end{equation}
where $\mathcal{P}_{Cali}^{-1}\in \mathcal{R}^{4x4}$ and $\mathcal{P}_{ECEF}\in \mathcal{R}^{4x4}$ are the inverse of the LiDAR calibration matrix, and the ECEF transformation matrix, respectively. $X_{hor}$, $Y_{hor}$, and $Z_{hor}$ represent the coordinates of 3D points concerning Horizontal LiDAR Coordinate. The $\mathcal{P}_{ECEF}$ matrix responsible for transforming points in LiDAR coordinate frame to the geocentric coordinate frame (ECEF) is calculated using the Ground Control Point surveying technique~\cite{Schuhmacher2005GeoreferencingOT}. 

The longitude $(\lambda)$ is calculated from the ECEF position using Eq.~\ref{eq:latlon},

\begin{equation}
\label{eq:latlon}
    \lambda = \arctan(\frac{Y_{ecef}}{X_{ecef}})
\end{equation}

The geodetic latitude $(\phi)$ is calculated using Bowring's method by solving Eq.~\ref{eq:reduced_latitude} and Eq.~\ref{eq:geo_latitude} in an iterative manner,

\begin{equation}
\label{eq:reduced_latitude}
   \overline{\beta} = \arctan(\frac{Z_{ecef}}{(1-f)s})
\end{equation}

\begin{equation}
\label{eq:geo_latitude}
   \overline{\phi} = \arctan(\dfrac{Z_{ecef} + e^2(\dfrac{1-f}{1-e^2})R(\sin{\beta})^3}{s - e^2R(\cos{\beta})^3})
\end{equation}

where $R$, $f$, and $e^2 = 1-(1-f)^2$ are the equatorial radius, flattening of the planet, and the square of first eccentricity, respectively. $s$ is defined as $s = \sqrt{X_{ecef}^2 + Y_{ecef}^2}$. The altitude ($h_{ego}$, height above ellipsoid) is given by,

\begin{equation}
    h_{ego} = s\cos{\phi} + (Z_{ecef} + e^2N\sin{\phi})\sin{\phi} - N
\end{equation}

where $N$, the radius of curvature in the vertical prime, is defined as 

\begin{equation}
N = \dfrac{R}{\sqrt{1-e^2(\sin{\phi})^2}}
\end{equation}

Then the geo-referenced perception information ($\phi, \lambda, h_{ego}$) along with other data will be transmitted to the cloud server for distribution and the final data is packaged as:
\begin{equation}
Data_{roadside} = \{M^{(i)}(t, id, \phi, \lambda, h_{ego}, w, l, h, \theta) \}_{i= 1}^{N_{Dbbx}}
\end{equation}


\subsection{Cloud Communication}

\begin{figure}[!h]
    \centering
    \includegraphics[width=0.5\textwidth]{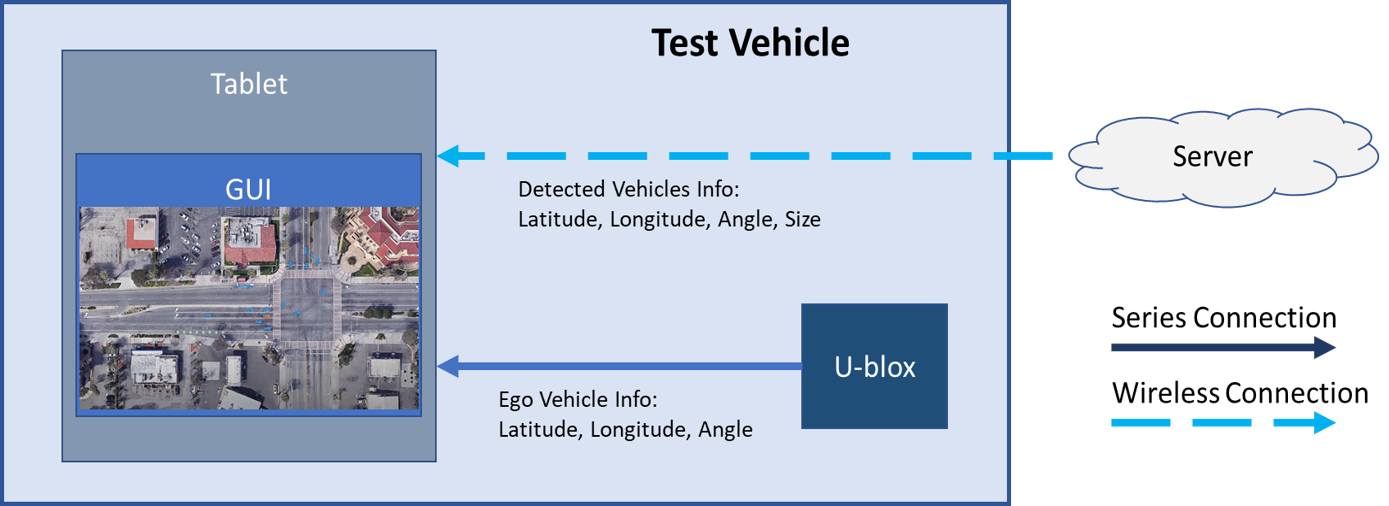}
    \caption{Illustration of onboard settings (structure and communications).}
    \label{fig:CMM_OBU}
\end{figure}

As shown in Fig.~\ref{fig:CMM_OBU}, the onboard unit (OBU) retrieves traffic perception data from the cloud server and GPS location data from a GPS receiver. Then the onboard unit reconstructs the traffic conditions based on the multi-source data and displays it on the graphical user interface (GUI) in real-time (the update frequency is 3-4 Hz on average). In our field implementation, a Samsung Galaxy Tab A7 tablet serves as an OBU, running a designed application to retrieve data from the GPS receiver and displaying the reconstructed object-level traffic information on the GUI. We adopt a NETGEAR AirCard 770S mobile hotspot which is equipped with a 4G/LTE sim card and can provide Vehicle-to-Cloud (V2C) communication between the cloud server and OBU. To have accurate GPS measurements, we utilize a C102-F9R U-Blox unit with an embedded Inertial Measurement Unit (IMU) which provides an 8Hz update frequency on the GPS location and heading.

\subsection{Multi-Object Reconstruction}

An application is designed to visualize the location of vehicles perceived by the roadside unit (RSU) and the ego vehicle provided by the OBU. To achieve that, we first locate the monitored area at the intersection and crop it from the Google Earth Pro satellite view. We leverage the cropped image as a background map for visualizing the reconstructed traffic. Firstly, we calculate the distance between two reference GPS points using the Haversine formula as shown followed. 
\begin{equation}
\begin{aligned}
a =& sin^{2}(\Delta lat/2) + cos(lat_{ref1}) \\
&\cdot cos(lat_{ref2})\cdot sin^{2}(\Delta lon/2)\\    
c =& 2\cdot atan2( \sqrt{a}, \sqrt{1-a} )\\
d =& R\cdot c
\end{aligned}
\end{equation}
where $lat_{ref1}$ and $lat_{ref2}$ are latitudes of two reference GPS points, $\Delta lat$ is the latitude difference between two GPS points, $\Delta lon$ is the longitude difference between two GPS points, $R$ is the radius of the earth, and $d$ is the distance computed between two GPS points.
Based on the number of pixels between their displayed pixel coordinates on the tablet, we can calculate the transfer ratio between them. 
\begin{equation}
\frac{Pix_{ref1} - Pix_{ref2}}{Dis_{ref1} - Dis_{ref2}} = \alpha
\end{equation}
where, $Pix_{ref1}$ and $Pix_{ref2}$ are the pixel coordinates of two reference points,  $Dis_{ref1}$ and  $Dis_{ref2}$ are the distance between two reference points, and $\alpha$ is the transfer ratio.
By now, we can create an object and display it on the desired pixel coordinates based on its GPS location.


The tablet and the u-blox are wire-connected and the data is transmitted via Universal Serial Bus (USB) serial connection between them. With the GPS location and heading, the ego vehicle is displayed on the GUI as an orange vehicle icon. On the other hand, the data from the cloud server contains the perception information, including GPS location, heading, and size three-dimension, obtained from the RSU. From the cloud server data, we first separate the vehicle data from the pedestrian data based on the three-dimensional size information. Then display the vehicles sensed by the RSU with blue vehicle icons and pedestrians with pedestrian top view icons.

\section{Field Testing and Results Analysis}
\label{cmm results}
\subsection{Feasibility}

Object-level perception information acts as the building block for CMM, which requires high-fidelity data retrieved from high-resolution sensors, such as LiDARs. Nevertheless, it could be costly, time-consuming, and to some extent, restricted by policies and protocols, to deploy these sensors directly in the real world. Thus, it is necessary to evaluate the feasibility of the system at the early stage of this work.
\begin{figure}[!h]
    \centering
    \includegraphics[width=0.5\textwidth]{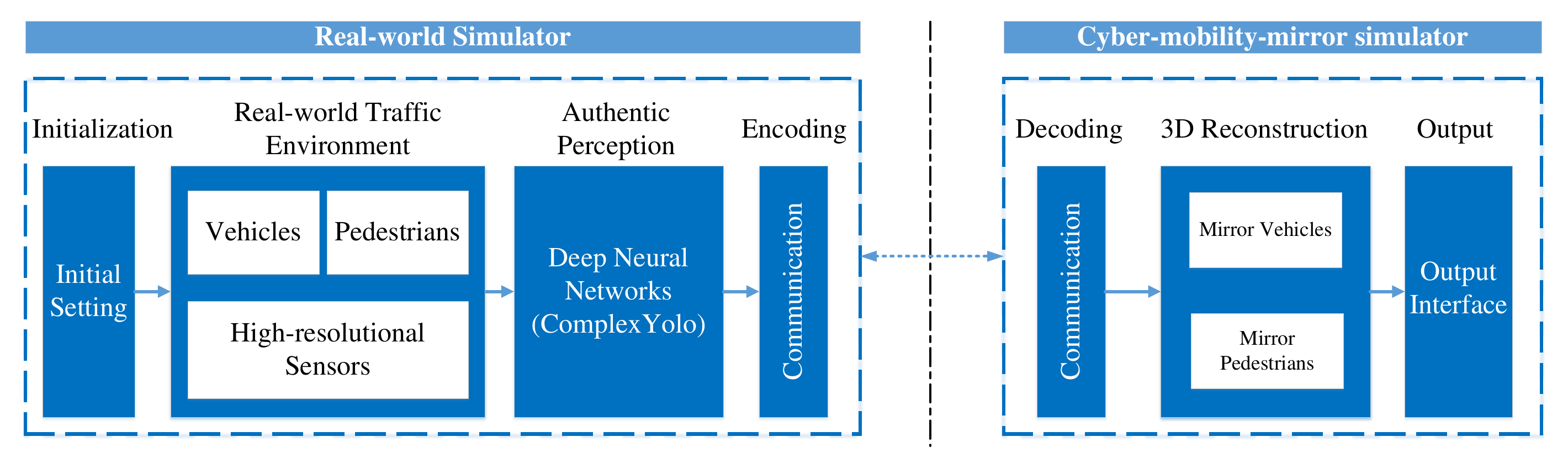}
    \caption{Structure for the CMM-based co-simulation platform.}
    \label{fig:co-simulation}
\end{figure}
    
To find an efficient and cost-effective way to validate the feasibility of CMM, we emulated a CMM system in a simulation platform, i.e., a CARLA-based co-simulation system~\cite{bai2022cyber}, before the real-world implementation. As demonstrated in Fig.~\ref{fig:co-simulation}, the basic idea is to emulate the real-world traffic environment via one CARLA simulator~\cite{dosovitskiy2017carla} and run the entire perception process within the emulated real-world environment. Then the other CARLA simulator is applied to emulate the cyber world, i.e., to reconstruct the traffic objects and then display them. Owing to the capability of CARLA to model high-fidelity sensors, the evaluation results of the emulated CMM in the co-simulation platform can lay the foundation for real-world CMM implementation.

\begin{figure}[!ht]
    \centering
    \includegraphics[width=0.5\textwidth]{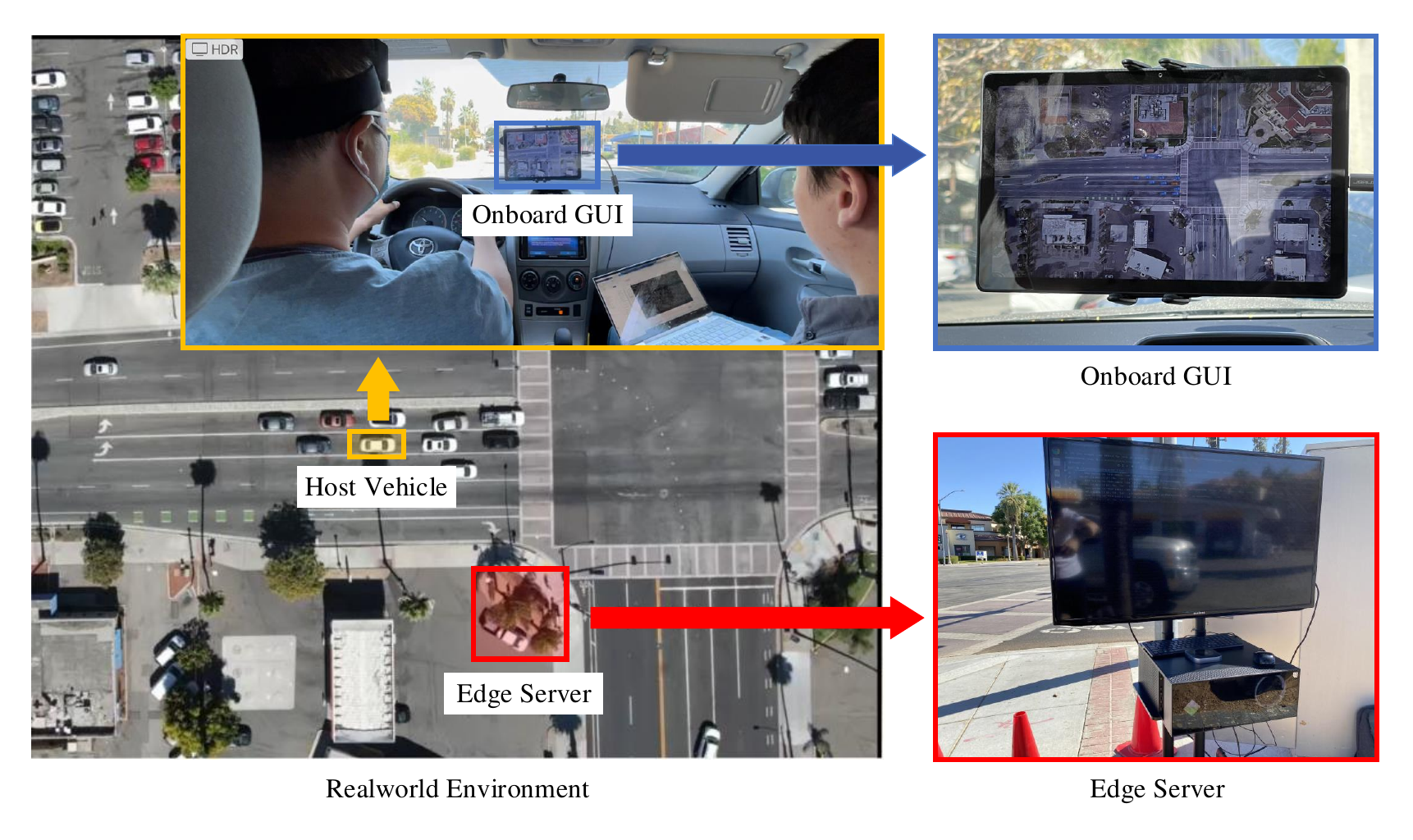}
    \caption{Illustration of CMM field operational test from different views from a drone, host vehicle, onboard GUI, and edge server.}
    \label{fig:CMM_Result_All}
\end{figure}

After the feasibility check in the simulation environment, we implement the CMM field operational system (FOS) at a real-world intersection of University Ave. \& Iowa Ave. in Riverside, California. Fig.~\ref{fig:CMM_Result_All} depicts the field system from different views. Multi-view videos are captured along the test including drone's view, in-vehicle views (including  driver perspective, backseat passenger perspective, and GUI), roadside view, and point cloud data-based bird's-eye view (BEV). A video clip is edited with the descriptive annotations to show the whole online process, which is available at \url{https://www.youtube.com/watch?v=0egpmgkzyG0}). The video demonstrates the feasibility of the CMM FOS and the following sections will show the results of detection accuracy and real-time performance.

\subsection{Detection}

\begin{figure*}[!ht]
    \centering
    \includegraphics[width=\textwidth]{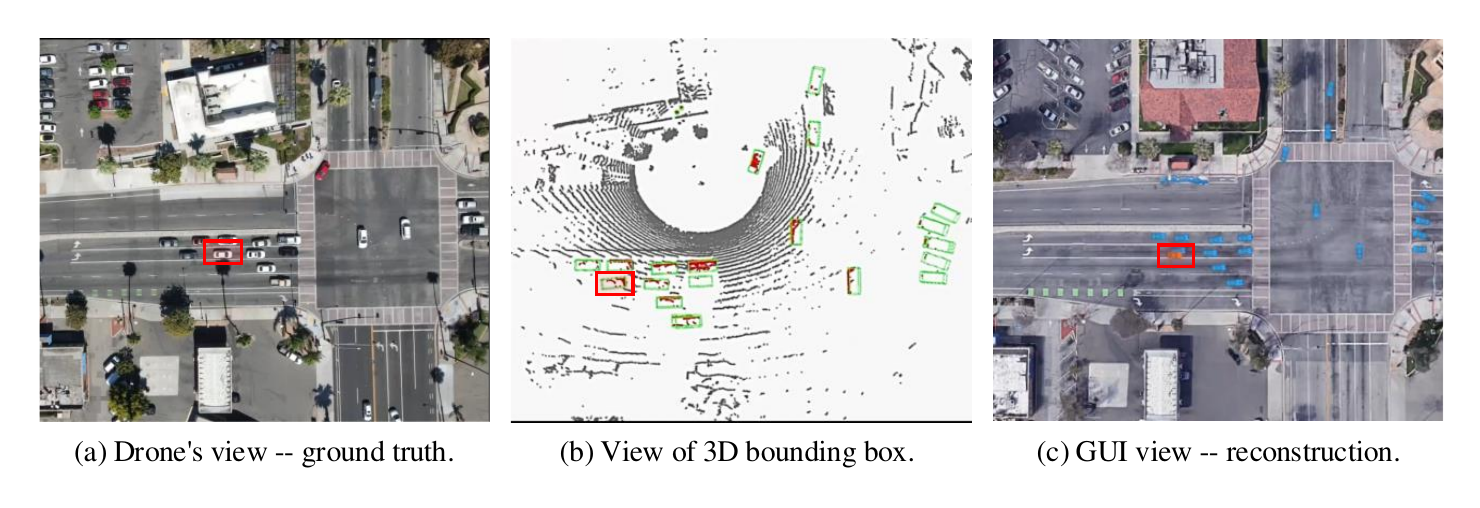}
    \caption{Examples of the CMM FOS testing results from different perspectives (The ego-vehicle is marked by red boxes).}
    \label{fig:resutls_examples}
\end{figure*}
Fig.~\ref{fig:resutls_examples} demonstrates several frames of the CMM FOS testing results. The first column shows the drone view, the second column depicts the bird's-eye view from LiDAR data, and the third column presents the reconstructed view on the onboard graphical user interface. The ego vehicle equipped with the CMM onboard system is marked by a red rectangle in each figure. In the GUI, the orange icons represent the GPS locations of the ego vehicle, while the blue ones denote vehicles detected by the roadside LiDAR. Additionally, pedestrians are also detected and shown in the GUI with top view pedestrian icons (shown in the video). 
The detection accuracy is evaluated by the \textit{Confusion Matrix}, a popular evaluation process used in the computer vision area~\cite{everingham2015pascal}.

Specifically, the detection results can be  categorized into four classes:
\begin{itemize}
    \item \textbf{True Positive (TP)}: the number of cases predicted as positive by the classifier when they are indeed positive, i.e., a vehicle object is detected as a vehicle.
    
    \item \textbf{False Positive (FP)} = the number of cases predicted as positive by the classifier when they are indeed negative, i.e., a non-vehicle object is detected as a vehicle.
    
    \item \textbf{True Negative (TN)} = the number of cases predicted as negative by the classifier when they are indeed negative, i.e., a non-vehicle object is detected as a non-vehicle object.

    \item \textbf{False Negative (FN)} = the number of cases predicted as negative by the classifier when they are indeed positive, i.e., a vehicle is detected as a non-vehicle object.
\end{itemize}

\textit{Precision} is the ability of the detector to identify only relevant objects, i.e., vehicles and pedestrians in this paper. It is the proportion of correct positive predictions and is given by
\begin{equation}
    Precision = \frac{TP}{TP+FP} = \frac{TP}{\text{\# of all detections}}
\end{equation}
\textit{Recall} is a metric that measures the ability of the detector to find all the relevant cases (that is, all the ground truths). It is the proportion of true positive detected among all ground-truth (i.e., real vehicles) and is defined as
\begin{equation}
    Recall = \frac{TP}{TP+FN} = \frac{TP}{\text{\# of all ground truth}}
\end{equation}

In terms of the perspective for traffic surveillance, we define another metric named \textit{Miss} which measures the portion of ``missing'' vehicles (that are not detected) and is defined by
\begin{equation}
    Miss = \frac{FN}{TP+TN} = \frac{\text{\# of all missing vehicles}}{\text{\# of all ground truth}}
\end{equation}

To evaluate the prototype system performance, we randomly select 130 frames of testing data and manually label them based on the drone's view. A total of 1661 vehicles are labeled as the ground truth and the detection accuracy is evaluated based on the three aforementioned parameters. Table~\ref{tab:performance} summarizes the evaluation results.

\begin{table}[!h]
    \centering
    \caption{Test performance of CMM FOS}
    \label{tab:performance}
    \begin{tabular}{c|c|c|c|c|c}
    \hline
        Ground Truth & TP & FP & Precision& Recall  & Miss \\\hline
        1661 & 1389 & 43 & 96.99\%& 83.62\%  & 16.38\%\\
    \hline
    \end{tabular}
\end{table}

\subsection{Localization}
This section analyzes the localization performance of our CMM field operational system. To evaluate the localization accuracy, a multi-sensor-based localization system is applied to measure the ground truth location of the ego-vehicle. This multi-sensor system consists of a GPS receiver enabled with Real-Time Kinematic (RTK) positioning and an Inertial Measurement Unit (IMU). Since this system can achieve centimeter-level positioning, the measurement generated by this GPS-RTK-IMU positioning system is used as the ground truth to assess the CMM system.

\begin{figure*}[!ht]
    \centering
    \includegraphics[width=\textwidth]{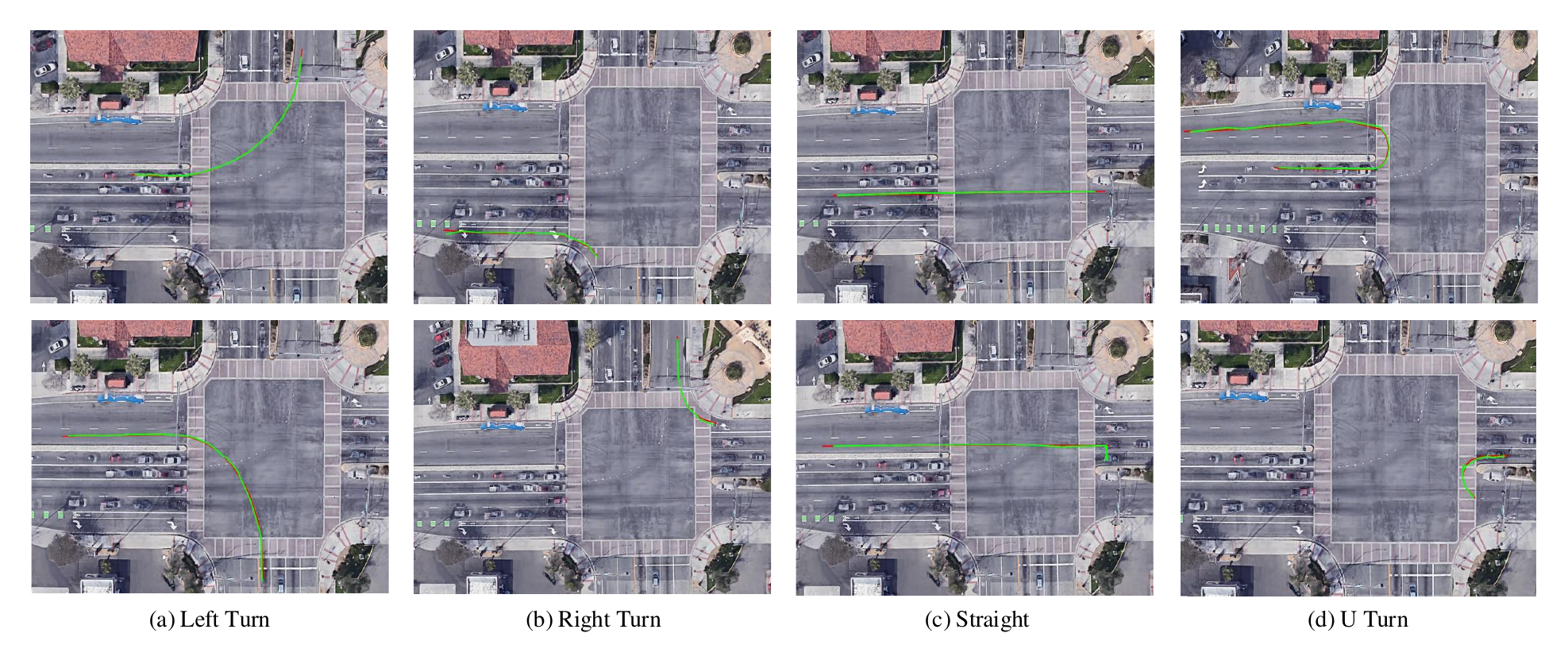}
    \caption{Trajectories of different driving scenarios (the trajectories from CMM FOS and ground truth are shown in green and red, respectively).}
    \label{fig:CMM_localization}
\end{figure*}

Field tests are conducted in terms of different driving scenarios, including 1) left turn, 2) right turn, 3) going straight, and 4) U-turn. The trajectories of ego-vehicle with four driving scenarios are extracted and visualized in Fig.~\ref{fig:CMM_localization}. From the subfigures shown in Fig.~\ref{fig:CMM_localization}, the trajectories generated by our CMM system (green curves) highly match the ground truth generated by the onboard GPS-RTK-IMU positioning system (red curves). 

\begin{figure}[!ht]
    \centering
    \subfigure[Histogram of the localization error.]
    {\includegraphics[width=0.5\textwidth]{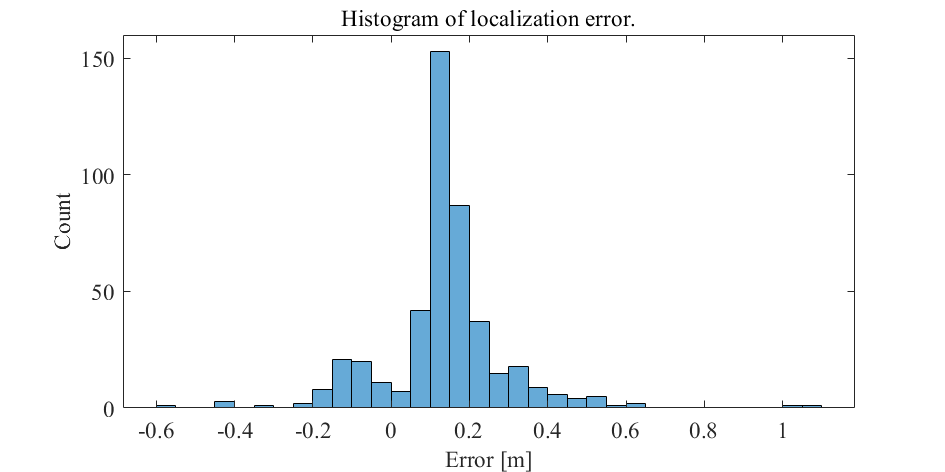}
    \label{fig:loc-err-a}}
    \subfigure[Boxplot of the localization error.]
    {\includegraphics[width=0.5\textwidth]{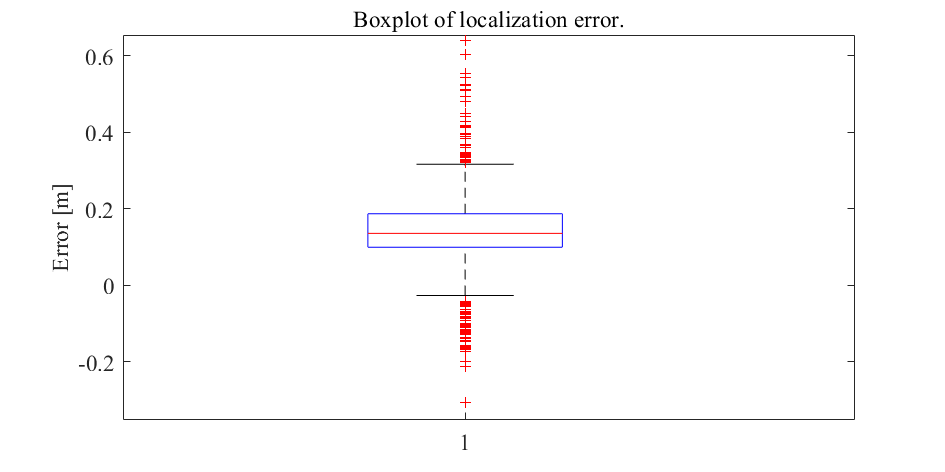}
    \label{fig:loc-err-b}}
    \caption{Localization error analysis between CMM and ground truth.}
    \label{fig:loc-err}
\end{figure}

Fig.~\ref{fig:loc-err} shows the quantitative analysis results. Totally 455 frames of data are selected in terms of different driving scenarios and according to Fig.~\ref{fig:loc-err-a}, most of the errors (52.7\%) fall into the interval of $[0.1m, 0.2m]$. Additionally, 62.5\% localization results have errors within $[-0.2m, 0.2m]$. From the boxplot analysis in Fig.~\ref{fig:loc-err-b}, excepting the outliers, the minimum and maximum localization errors are $-0.03m$ and $0.32m$, respectively, which ensures the applicability of our system for CDA applications in the real-world traffic environment.

\subsection{Latency}
\begin{figure*}[!ht]
    \centering
    \includegraphics[width=\textwidth]{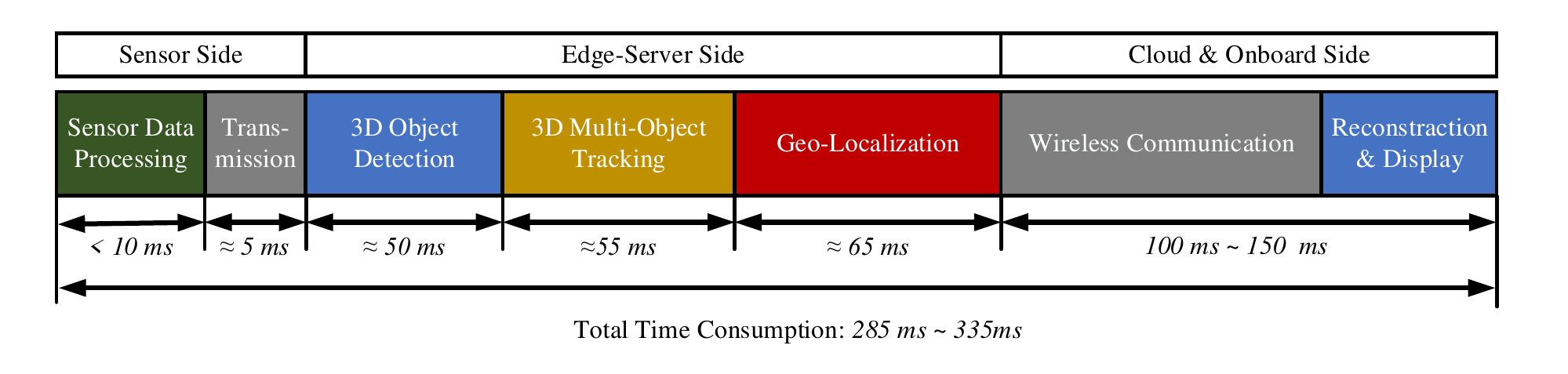}
    \caption{Visualization of the latency at different stages in CMM FOS.}
    \label{fig:CMM_latency}
\end{figure*}

As for a field operation system (FOS), it is of great significance to analyze the latency of the whole system. Depicted in Fig.~\ref{fig:CMM_latency}, the latency of the whole CMM FOS pipeline can be analyzed by breaking down the whole workflow into three main phases:

\begin{itemize}
    \item \textbf{Phase 1 -- Sensor Side}: Time elapsed from the start till the edge server receives the sensor data. Specifically, in the \textit{sensor processing} stage, the sensor collects the raw data and processes it into a transformable format via its embedded system. For data retrieving, the processed data can be transmitted to the edge server via the Local Area Network (LAN). The time consumption is certified by the manufacturer. 
    
    \item \textbf{Phase 2 -- Edge-Server Side}: Time elapsed from the moment when sensor data is received by the edge server till the instance when perception data is encoded and sent out to the cloud server. The edge server is responsible for generating the object-level perception data, including 3D object detection, 3D multi-object tracking, and geodetic localization. Since these modules are running in chronological order, the time consumption for each module is measured by the starting and ending timestamps of each function. 
    
    \item \textbf{Phase 3 -- Cloud \& Onboard}: Time elapsed from the moment when perception data is sent from the edge server till the instance when reconstructed traffic environments are displayed on the onboard GUI. Since the CMM system tends to serve all the road users with connectivity, a cloud server is used for data acquisition, synchronization, and distribution of processed data (after edge computing). The onboard computer, i.e., the tablet utilized in this study, decodes the perception data, reconstructs the traffic environment, and displays it on the GUI. Time consumption for this phase is measured by the timestamps from the onboard end to the edge-server end.
\end{itemize}

As shown in Fig.~\ref{fig:CMM_latency}, the total latency is about $285ms - 335ms$, whose variance mainly results from the fluctuation of communication. However, during the field testing, we find out that the time consumption of every single computational module may vary within a certain range. For example, the object-tracking and geo-localization modules have a larger variance compared with the object detection model, which may be caused by the change in the number of detected objects. 

To reduce the latency of the whole system, there are several ways that can be applied in the future. For example, several \textit{for loops} and \textit{external python packages} are implemented in the software for tracking and localization parts which mainly account for the surprisingly high computational cost at the perception end. Therefore, programming optimization can be applied to further reduce computational time. Another way to speed up the whole process is to improve the hardware's computational performance for edge servers and onboard computers.

\section{Conclusion and Discussion}
In this study, we introduce the concept of Cyber Mobility Mirror (CMM) and develop a CMM Field Operational System at a real-world intersection as a prototype for enabling Cooperative Driving Automation (CDA). It leverages high-fidelity roadside sensors (e.g., LiDAR) to detect, classify, track and reconstruct object-level traffic information in real-time, which can lay a foundation of environment perception for various kinds of CDA applications in mixed traffic. Testing results prove the feasibility of the CMM concept and also demonstrate satisfactory system performance in terms of real-time high-fidelity traffic surveillance. The overall perception accuracy metrics include 96.99\% for precision and 83.62\% for recall. Additionally, the average geo-localization error of the system is $0.14m$ and real-time traffic conditions can be displayed at a frequency of $3-4 Hz$.


Based on this prototype CMM FOS, several future directions for improving the system performance may include: 
\begin{itemize}
    \item \textit{\textbf{Perception Accuracy}}: Since it is a cost-effective way to collect roadside training datasets from the SOTA autonomous driving simulators, e.g., CARLA~\cite{dosovitskiy2017carla}, we will improve the detection accuracy by enhancing the model with transferability, i.e., training on simulation and testing on real-world; 
    \item \textit{\textbf{Perception Range}}: The current CMM FOS only involves one LiDAR sensor and thus can only cover a limited area of the whole intersection. To extend the perception range of the CMM system, we plan to set up several sensors including both LiDARs and cameras to cover multiple intersections to achieve a corridor-level cooperative perception system;
    \item \textit{\textbf{Real-time Performance}}: The time consumption can be mainly reduced from the edge-server side, i.e., optimizing the software programming in the tracking and localization parts. Besides, upgrading the hardware equipment can also improve the real-time processing speed. 
\end{itemize}
    
This paper intends to provide a field operational system of a novel concept of the roadside sensor-based high-fidelity traffic surveillance system, named CMM, which hopes can provide foundations and inspirations for future work. By leveraging the high-fidelity roadside sensing information available from the CMM system, plenty of subsequent CDA applications (e.g., CACC, advanced intersection management, cooperative eco-driving) can be revisited for real-world implementation in the mixed traffic environment. 

\section*{Acknowledgment}

This research was funded by the Toyota Motor North America InfoTech Labs. The contents of this paper reflect the views of the authors, who are responsible for the facts and the accuracy of the data presented herein. The contents do not necessarily reflect the official views of Toyota Motor North America.

\ifCLASSOPTIONcaptionsoff
  \newpage
\fi



%
\bibliographystyle{IEEEtran}
\bibliography{references}{}

\begin{thebibliography}{10}
\providecommand{\url}[1]{#1}
\csname url@samestyle\endcsname
\providecommand{\newblock}{\relax}
\providecommand{\bibinfo}[2]{#2}
\providecommand{\BIBentrySTDinterwordspacing}{\spaceskip=0pt\relax}
\providecommand{\BIBentryALTinterwordstretchfactor}{4}
\providecommand{\BIBentryALTinterwordspacing}{\spaceskip=\fontdimen2\font plus
\BIBentryALTinterwordstretchfactor\fontdimen3\font minus
  \fontdimen4\font\relax}
\providecommand{\BIBforeignlanguage}[2]{{%
\expandafter\ifx\csname l@#1\endcsname\relax
\typeout{** WARNING: IEEEtran.bst: No hyphenation pattern has been}%
\typeout{** loaded for the language `#1'. Using the pattern for}%
\typeout{** the default language instead.}%
\else
\language=\csname l@#1\endcsname
\fi
#2}}
\providecommand{\BIBdecl}{\relax}
\BIBdecl

\bibitem{fagnant2015preparing}
D.~J. Fagnant and K.~Kockelman, ``Preparing a nation for autonomous vehicles:
  opportunities, barriers and policy recommendations,'' \emph{Transportation
  Research Part A: Policy and Practice}, vol.~77, pp. 167--181, 2015.

\bibitem{misener2006path}
J.~A. Misener and S.~E. Shladover, ``Path investigations in vehicle-roadside
  cooperation and safety: A foundation for safety and vehicle-infrastructure
  integration research,'' in \emph{2006 IEEE Intelligent Transportation Systems
  Conference}.\hskip 1em plus 0.5em minus 0.4em\relax IEEE, 2006, pp. 9--16.

\bibitem{stahlmann2011starting}
R.~Stahlmann, A.~Festag, A.~Tomatis, I.~Radusch, and F.~Fischer, ``Starting
  european field tests for car-2-x communication: the drive c2x framework,'' in
  \emph{18th ITS World Congress and Exhibition}, 2011, p.~12.

\bibitem{carma}
\BIBentryALTinterwordspacing
USDOT, ``Carma program overview,'' May 2021. [Online]. Available:
  \url{https://highways.dot.gov/research/operations/CARMA}
\BIBentrySTDinterwordspacing

\bibitem{eucar_2021}
\BIBentryALTinterwordspacing
EUCAR, ``Autonet2030,'' May 2021. [Online]. Available:
  \url{https://www.autonet2030.eu/}
\BIBentrySTDinterwordspacing

\bibitem{arnold2019survey}
E.~Arnold, O.~Y. Al-Jarrah, M.~Dianati, S.~Fallah, D.~Oxtoby, and
  A.~Mouzakitis, ``A survey on 3d object detection methods for autonomous
  driving applications,'' \emph{IEEE Transactions on Intelligent Transportation
  Systems}, vol.~20, no.~10, pp. 3782--3795, 2019.

\bibitem{yurtsever2020survey}
E.~Yurtsever, J.~Lambert, A.~Carballo, and K.~Takeda, ``A survey of autonomous
  driving: Common practices and emerging technologies,'' \emph{IEEE access},
  vol.~8, pp. 58\,443--58\,469, 2020.

\bibitem{bai2022infrastructure}
Z.~Bai, G.~Wu, X.~Qi, Y.~Liu, K.~Oguchi, and M.~J. Barth,
  ``Infrastructure-based object detection and tracking for cooperative driving
  automation: A survey,'' \emph{arXiv preprint arXiv:2201.11871}, 2022.

\bibitem{datondji2016survey}
S.~R.~E. Datondji, Y.~Dupuis, P.~Subirats, and P.~Vasseur, ``A survey of
  vision-based traffic monitoring of road intersections,'' \emph{IEEE
  transactions on intelligent transportation systems}, vol.~17, no.~10, pp.
  2681--2698, 2016.

\bibitem{wu2020automatic}
J.~Wu, H.~Xu, J.~Zheng, and J.~Zhao, ``Automatic vehicle detection with
  roadside lidar data under rainy and snowy conditions,'' \emph{IEEE
  Intelligent Transportation Systems Magazine}, vol.~13, no.~1, pp. 197--209,
  2020.

\bibitem{zhao2019detection}
J.~Zhao, H.~Xu, H.~Liu, J.~Wu, Y.~Zheng, and D.~Wu, ``Detection and tracking of
  pedestrians and vehicles using roadside lidar sensors,'' \emph{Transportation
  research part C: emerging technologies}, vol. 100, pp. 68--87, 2019.

\bibitem{wu2018automatic}
J.~Wu, H.~Xu, and J.~Zhao, ``Automatic lane identification using the roadside
  lidar sensors,'' \emph{IEEE Intelligent Transportation Systems Magazine},
  vol.~12, no.~1, pp. 25--34, 2018.

\bibitem{wu2020improved}
J.~Wu, H.~Xu, Y.~Zhang, and R.~Sun, ``An improved vehicle-pedestrian near-crash
  identification method with a roadside lidar sensor,'' \emph{Journal of safety
  research}, vol.~73, pp. 211--224, 2020.

\bibitem{zhang2019automatic}
Z.~Zhang, J.~Zheng, H.~Xu, X.~Wang, X.~Fan, and R.~Chen, ``Automatic background
  construction and object detection based on roadside lidar,'' \emph{IEEE
  Transactions on Intelligent Transportation Systems}, vol.~21, no.~10, pp.
  4086--4097, 2019.

\bibitem{lv2019lidar}
B.~Lv, H.~Xu, J.~Wu, Y.~Tian, Y.~Zhang, Y.~Zheng, C.~Yuan, and S.~Tian,
  ``Lidar-enhanced connected infrastructures sensing and broadcasting
  high-resolution traffic information serving smart cities,'' \emph{IEEE
  Access}, vol.~7, pp. 79\,895--79\,907, 2019.

\bibitem{song2020background}
Y.~Song, H.~Zhang, Y.~Liu, J.~Liu, H.~Zhang, and X.~Song, ``Background
  filtering and object detection with a stationary lidar using a layer-based
  method,'' \emph{IEEE Access}, vol.~8, pp. 184\,426--184\,436, 2020.

\bibitem{ester1996density}
M.~Ester, H.-P. Kriegel, J.~Sander, X.~Xu \emph{et~al.}, ``A density-based
  algorithm for discovering clusters in large spatial databases with noise.''
  in \emph{kdd}, vol.~96, no.~34, 1996, pp. 226--231.

\bibitem{zou2019object}
Z.~Zou, Z.~Shi, Y.~Guo, and J.~Ye, ``Object detection in 20 years: A survey,''
  \emph{arXiv preprint arXiv:1905.05055}, 2019.

\bibitem{cucchiara2000statistic}
R.~Cucchiara, C.~Grana, M.~Piccardi, and A.~Prati, ``Statistic and
  knowledge-based moving object detection in traffic scenes,'' in
  \emph{ITSC2000. 2000 IEEE Intelligent Transportation Systems. Proceedings
  (Cat. No. 00TH8493)}.\hskip 1em plus 0.5em minus 0.4em\relax IEEE, 2000, pp.
  27--32.

\bibitem{aslani2013optical}
S.~Aslani and H.~Mahdavi-Nasab, ``Optical flow based moving object detection
  and tracking for traffic surveillance,'' \emph{International Journal of
  Electrical, Computer, Energetic, Electronic and Communication Engineering},
  vol.~7, no.~9, pp. 1252--1256, 2013.

\bibitem{huang2013highly}
S.-C. Huang and B.-H. Chen, ``Highly accurate moving object detection in
  variable bit rate video-based traffic monitoring systems,'' \emph{IEEE
  transactions on neural networks and learning systems}, vol.~24, no.~12, pp.
  1920--1931, 2013.

\bibitem{boukerche2021object}
A.~Boukerche and Z.~Hou, ``Object detection using deep learning methods in
  traffic scenarios,'' \emph{ACM Computing Surveys (CSUR)}, vol.~54, no.~2, pp.
  1--35, 2021.

\bibitem{redmon2016you}
J.~Redmon, S.~Divvala, R.~Girshick, and A.~Farhadi, ``You only look once:
  Unified, real-time object detection,'' in \emph{Proceedings of the IEEE
  conference on computer vision and pattern recognition}, 2016, pp. 779--788.

\bibitem{mao2020finding}
Q.-C. Mao, H.-M. Sun, L.-Q. Zuo, and R.-S. Jia, ``Finding every car: a traffic
  surveillance multi-scale vehicle object detection method,'' \emph{Applied
  Intelligence}, vol.~50, no.~10, pp. 3125--3136, 2020.

\bibitem{liu2016ssd}
W.~Liu, D.~Anguelov, D.~Erhan, C.~Szegedy, S.~Reed, C.-Y. Fu, and A.~C. Berg,
  ``Ssd: Single shot multibox detector,'' in \emph{European conference on
  computer vision}.\hskip 1em plus 0.5em minus 0.4em\relax Springer, 2016, pp.
  21--37.

\bibitem{wang2018multi}
X.~Wang, X.~Hua, F.~Xiao, Y.~Li, X.~Hu, and P.~Sun, ``Multi-object detection in
  traffic scenes based on improved ssd,'' \emph{Electronics}, vol.~7, no.~11,
  p. 302, 2018.

\bibitem{ren2015faster}
S.~Ren, K.~He, R.~Girshick, and J.~Sun, ``Faster r-cnn: Towards real-time
  object detection with region proposal networks,'' \emph{Advances in neural
  information processing systems}, vol.~28, pp. 91--99, 2015.

\bibitem{li2021method}
C.-j. Li, Z.~Qu, S.-y. Wang, and L.~Liu, ``A method of cross-layer fusion
  multi-object detection and recognition based on improved faster r-cnn model
  in complex traffic environment,'' \emph{Pattern Recognition Letters}, vol.
  145, pp. 127--134, 2021.

\bibitem{mhalla2018embedded}
A.~Mhalla, T.~Chateau, S.~Gazzah, and N.~E.~B. Amara, ``An embedded
  computer-vision system for multi-object detection in traffic surveillance,''
  \emph{IEEE Transactions on Intelligent Transportation Systems}, vol.~20,
  no.~11, pp. 4006--4018, 2018.

\bibitem{lian2021small}
J.~Lian, Y.~Yin, L.~Li, Z.~Wang, and Y.~Zhou, ``Small object detection in
  traffic scenes based on attention feature fusion,'' \emph{Sensors}, vol.~21,
  no.~9, p. 3031, 2021.

\bibitem{gahlert2020visibility}
N.~G{\"a}hlert, N.~Hanselmann, U.~Franke, and J.~Denzler, ``Visibility guided
  nms: Efficient boosting of amodal object detection in crowded traffic
  scenes,'' \emph{arXiv preprint arXiv:2006.08547}, 2020.

\bibitem{guindel2018fast}
C.~Guindel, D.~Martin, and J.~M. Armingol, ``Fast joint object detection and
  viewpoint estimation for traffic scene understanding,'' \emph{IEEE
  Intelligent Transportation Systems Magazine}, vol.~10, no.~4, pp. 74--86,
  2018.

\bibitem{zhang2020gc}
L.~Zhang, J.~Zheng, R.~Sun, and Y.~Tao, ``Gc-net: Gridding and clustering for
  traffic object detection with roadside lidar,'' \emph{IEEE Intelligent
  Systems}, 2020.

\bibitem{Liu9434525}
Z.~Liu, Q.~Li, S.~Mei, and M.~Huang, ``Background filtering and object
  detection with roadside lidar data,'' in \emph{2021 4th International
  Conference on Electron Device and Mechanical Engineering (ICEDME)}, 2021, pp.
  296--299.

\bibitem{redmond2007method}
S.~J. Redmond and C.~Heneghan, ``A method for initialising the k-means
  clustering algorithm using kd-trees,'' \emph{Pattern recognition letters},
  vol.~28, no.~8, pp. 965--973, 2007.

\bibitem{Song9216093}
Y.~Song, H.~Zhang, Y.~Liu, J.~Liu, H.~Zhang, and X.~Song, ``Background
  filtering and object detection with a stationary lidar using a layer-based
  method,'' \emph{IEEE Access}, vol.~8, pp. 184\,426--184\,436, 2020.

\bibitem{2021SAE}
SAE, ``Taxonomy and definitions for terms related to cooperative driving
  automation for on-road motor vehicles j3216\_202005,'' Available:
  \url{https://www.sae.org/standards/content/j3216_202005/}, 2021.

\bibitem{cuevas2005kalman}
E.~V. Cuevas, D.~Zaldivar, and R.~Rojas, ``Kalman filter for vision tracking,''
  2005.

\bibitem{okuma2004boosted}
K.~Okuma, A.~Taleghani, N.~De~Freitas, J.~J. Little, and D.~G. Lowe, ``A
  boosted particle filter: Multitarget detection and tracking,'' in
  \emph{European conference on computer vision}.\hskip 1em plus 0.5em minus
  0.4em\relax Springer, 2004, pp. 28--39.

\bibitem{bewley2016simple}
A.~Bewley, Z.~Ge, L.~Ott, F.~Ramos, and B.~Upcroft, ``Simple online and
  realtime tracking,'' in \emph{2016 IEEE international conference on image
  processing (ICIP)}.\hskip 1em plus 0.5em minus 0.4em\relax IEEE, 2016, pp.
  3464--3468.

\bibitem{8296962}
N.~c, A.~Bewley, and D.~Paulus, ``Simple online and realtime tracking with a
  deep association metric,'' in \emph{2017 IEEE International Conference on
  Image Processing (ICIP)}, 2017, pp. 3645--3649.

\bibitem{fernandez2019real}
M.~Fernandez-Sanjurjo, B.~Bosquet, M.~Mucientes, and V.~M. Brea, ``Real-time
  visual detection and tracking system for traffic monitoring,''
  \emph{Engineering Applications of Artificial Intelligence}, vol.~85, pp.
  410--420, 2019.

\bibitem{balamuralidhar2021multeye}
N.~Balamuralidhar, S.~Tilon, and F.~Nex, ``Multeye: Monitoring system for
  real-time vehicle detection, tracking and speed estimation from uav imagery
  on edge-computing platforms,'' \emph{Remote Sensing}, vol.~13, no.~4, p. 573,
  2021.

\bibitem{bolme2010visual}
D.~S. Bolme, J.~R. Beveridge, B.~A. Draper, and Y.~M. Lui, ``Visual object
  tracking using adaptive correlation filters,'' in \emph{2010 IEEE computer
  society conference on computer vision and pattern recognition}.\hskip 1em
  plus 0.5em minus 0.4em\relax IEEE, 2010, pp. 2544--2550.

\bibitem{chen2020edge}
C.~Chen, B.~Liu, S.~Wan, P.~Qiao, and Q.~Pei, ``An edge traffic flow detection
  scheme based on deep learning in an intelligent transportation system,''
  \emph{IEEE Transactions on Intelligent Transportation Systems}, vol.~22,
  no.~3, pp. 1840--1852, 2020.

\bibitem{cui2019automatic}
Y.~Cui, H.~Xu, J.~Wu, Y.~Sun, and J.~Zhao, ``Automatic vehicle tracking with
  roadside lidar data for the connected-vehicles system,'' \emph{IEEE
  Intelligent Systems}, vol.~34, no.~3, pp. 44--51, 2019.

\bibitem{kampker2018towards}
A.~Kampker, M.~Sefati, A.~S.~A. Rachman, K.~Kreisk{\"o}ther, and P.~Campoy,
  ``Towards multi-object detection and tracking in urban scenario under
  uncertainties.'' in \emph{VEHITS}, 2018, pp. 156--167.

\bibitem{zhang2020vehicle}
J.~Zhang, W.~Xiao, B.~Coifman, and J.~P. Mills, ``Vehicle tracking and speed
  estimation from roadside lidar,'' \emph{IEEE Journal of Selected Topics in
  Applied Earth Observations and Remote Sensing}, vol.~13, pp. 5597--5608,
  2020.

\bibitem{herrera2007traffic}
J.~C. Herrera and A.~M. Bayen, ``Traffic flow reconstruction using mobile
  sensors and loop detector data,'' 2007.

\bibitem{jiang2018compressive}
D.~Jiang, W.~Wang, L.~Shi, and H.~Song, ``A compressive sensing-based approach
  to end-to-end network traffic reconstruction,'' \emph{IEEE Transactions on
  Network Science and Engineering}, vol.~7, no.~1, pp. 507--519, 2018.

\bibitem{9107468}
M.~Cao, L.~Zheng, W.~Jia, and X.~Liu, ``Joint 3d reconstruction and object
  tracking for traffic video analysis under iov environment,'' \emph{IEEE
  Transactions on Intelligent Transportation Systems}, vol.~22, no.~6, pp.
  3577--3591, 2021.

\bibitem{9151364}
Q.~Rao and S.~Chakraborty, ``In-vehicle object-level 3d reconstruction of
  traffic scenes,'' \emph{IEEE Transactions on Intelligent Transportation
  Systems}, pp. 1--13, 2020.

\bibitem{bai2022hybrid}
Z.~Bai, P.~Hao, W.~Shangguan, B.~Cai, and M.~J. Barth, ``Hybrid reinforcement
  learning-based eco-driving strategy for connected and automated vehicles at
  signalized intersections,'' \emph{IEEE Transactions on Intelligent
  Transportation Systems}, pp. 1--14, 2022.

\bibitem{wangCACC}
Z.~Wang, Y.~Bian, S.~E. Shladover, G.~Wu, S.~E. Li, and M.~J. Barth, ``A survey
  on cooperative longitudinal motion control of multiple connected and
  automated vehicles,'' \emph{IEEE Intelligent Transportation Systems
  Magazine}, vol.~12, no.~1, pp. 4--24, 2020.

\bibitem{caesar2020nuscenes}
H.~Caesar, V.~Bankiti, A.~H. Lang, S.~Vora, V.~E. Liong, Q.~Xu, A.~Krishnan,
  Y.~Pan, G.~Baldan, and O.~Beijbom, ``nuscenes: A multimodal dataset for
  autonomous driving,'' in \emph{Proceedings of the IEEE/CVF conference on
  computer vision and pattern recognition}, 2020, pp. 11\,621--11\,631.

\bibitem{lang2019pointpillars}
A.~H. Lang, S.~Vora, H.~Caesar, L.~Zhou, J.~Yang, and O.~Beijbom,
  ``Pointpillars: Fast encoders for object detection from point clouds,'' in
  \emph{Proceedings of the IEEE/CVF Conference on Computer Vision and Pattern
  Recognition}, 2019, pp. 12\,697--12\,705.

\bibitem{yan2018second}
Y.~Yan, Y.~Mao, and B.~Li, ``Second: Sparsely embedded convolutional
  detection,'' \emph{Sensors}, vol.~18, no.~10, p. 3337, 2018.

\bibitem{lin2017focal}
T.-Y. Lin, P.~Goyal, R.~Girshick, K.~He, and P.~Doll{\'a}r, ``Focal loss for
  dense object detection,'' in \emph{Proceedings of the IEEE international
  conference on computer vision}, 2017, pp. 2980--2988.

\bibitem{Schuhmacher2005GeoreferencingOT}
S.~Schuhmacher and J.~Boehm, ``Georeferencing of terrestrial laserscanner data
  for applications in architectural modeling,'' 2005.

\bibitem{bai2022cyber}
Z.~Bai, G.~Wu, X.~Qi, K.~Oguchi, and M.~J. Barth, ``Cyber mobility mirror for
  enabling cooperative driving automation: A co-simulation platform,''
  \emph{arXiv preprint arXiv:2201.09463}, 2022.

\bibitem{dosovitskiy2017carla}
A.~Dosovitskiy, G.~Ros, F.~Codevilla, A.~Lopez, and V.~Koltun, ``Carla: An open
  urban driving simulator,'' in \emph{Conference on robot learning}.\hskip 1em
  plus 0.5em minus 0.4em\relax PMLR, 2017, pp. 1--16.

\bibitem{everingham2015pascal}
M.~Everingham, S.~Eslami, L.~Van~Gool, C.~K. Williams, J.~Winn, and
  A.~Zisserman, ``The pascal visual object classes challenge: A
  retrospective,'' \emph{International journal of computer vision}, vol. 111,
  no.~1, pp. 98--136, 2015.

\end{thebibliography}

%

\begin{IEEEbiography}
    [{\includegraphics[width=1in,height=1.25in,clip,keepaspectratio]{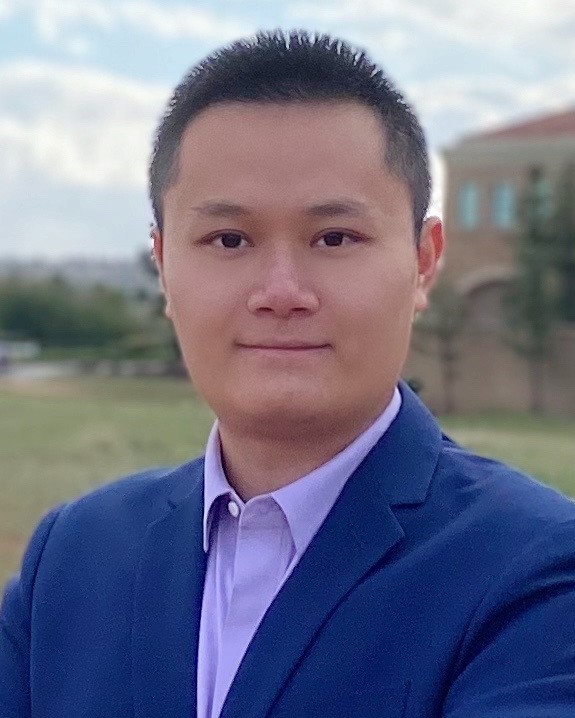}}]{Zhengwei Bai}
(Student Member, IEEE) received the B.E. and M.S. degrees from Beijing Jiaotong University, Beijing, China, in 2017 and 2020, respectively. He is currently a Ph.D. student in electrical and computer engineering at the University of California at Riverside. His research focuses on computer vision, sensor fusion, cooperative perception, and cooperative driving automation (CDA). He serves as a Review Editor in Urban Transportation Systems and Mobility.
\end{IEEEbiography}

\begin{IEEEbiography}
    [{\includegraphics[width=1in,height=1.25in,clip,keepaspectratio]{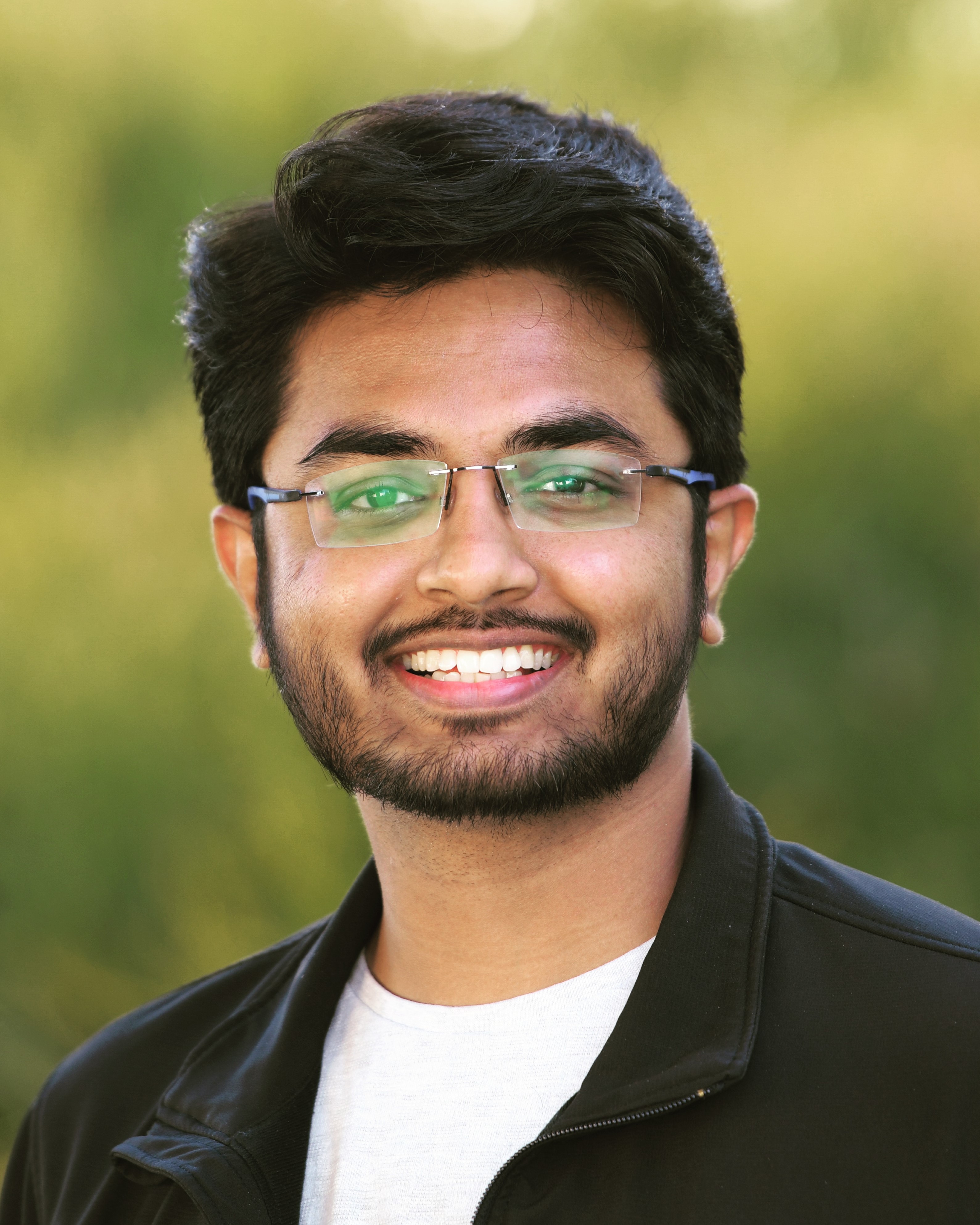}}]{Saswat N. Nayak}
received the B. Tech degree in Electrical Engineering from the National Institute of Technology Rourkela, India in 2018. He served as a Project Associate at the Department of Aerospace Engineering, Indian Institute of Technology Kanpur, India 2018-19. He is currently pursuing the Ph.D. degree at the Center of Environmental Research and Technology (CE-CERT), University of California Riverside, USA. His main research interests include vehicle positioning and localization in mixed traffic scenarios, multi-sensor fusion and connected vehicle applications.
\end{IEEEbiography}

\begin{IEEEbiography}
    [{\includegraphics[width=1in,height=1.25in,clip,keepaspectratio]{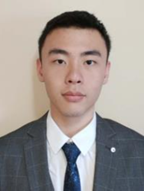}}]{Xuanpeng Zhao}
received the B.E. degree in electrical engineering from Shanghai Maritime University in 2019 and the M.S. degree in electrical engineering from the University of California at Riverside. He is currently a Ph.D. student in electrical and computer engineering at University of California at Riverside. His research focuses on cybersecurity, and connected and automated vehicle technology.
\end{IEEEbiography}

\begin{IEEEbiography}
    [{\includegraphics[width=1in,height=1.25in,clip,keepaspectratio]{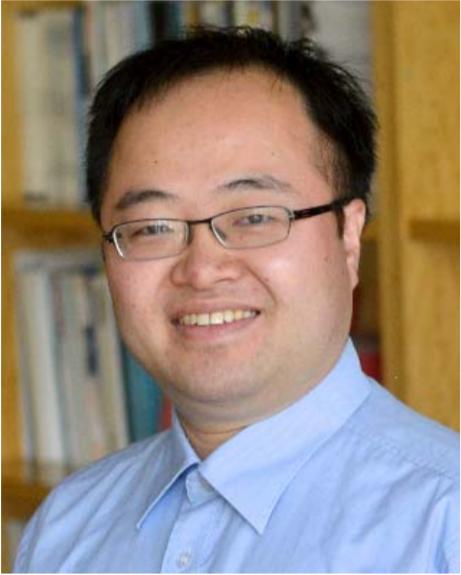}}]
    {Guoyuan Wu}
(Senior Member, IEEE) received his Ph.D. degree in mechanical engineering from the University of California, Berkeley in 2010. Currently, he holds an Associate Researcher and an Associate Adjunct Professor position at Bourns College of Engineering – Center for Environmental Research \& Technology (CE–CERT) and Department of Electrical \& Computer Engineering in the University of California at Riverside. development and evaluation of sustainable and intelligent transportation system (SITS) technologies, including connected and automated transportation systems (CATS), shared mobility, transportation electrification, optimization and control of vehicles, traffic simulation, and emissions measurement and modeling. Dr. Wu serves as Associate Editors for a few journals, including IEEE Transactions on Intelligent Transportation Systems, SAE International Journal of Connected and Automated Vehicles, and IEEE Open Journal of ITS. He is also a member of the Vehicle-Highway Automation Standing Committee (ACP30) of the Transportation Research Board (TRB), a board member of Chinese Institute of Engineers Southern California Chapter (CIE-SOCAL), and a member of Chinese Overseas Transportation Association (COTA). He is a recipient of Vincent Bendix Automotive Electronics Engineering Award.
\end{IEEEbiography}

\begin{IEEEbiography}
    [{\includegraphics[width=1in,height=1.25in,clip,keepaspectratio]{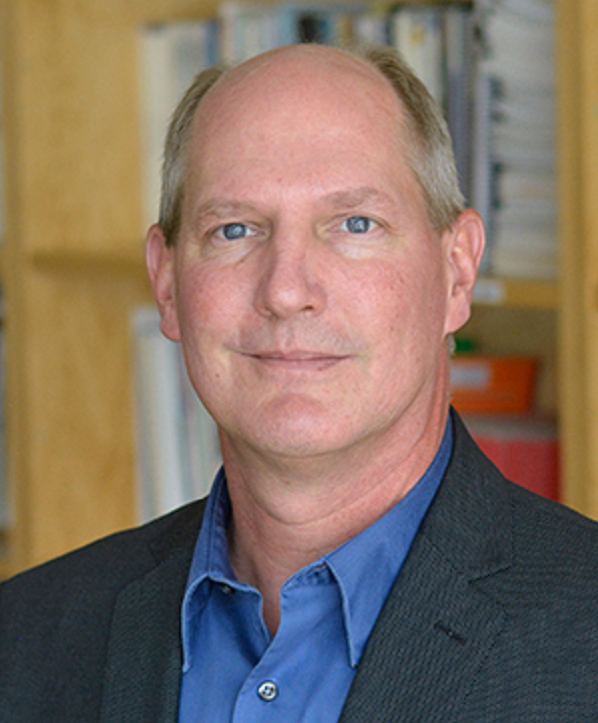}}]
    {Matthew J. Barth}
(Fellow, IEEE) received the M.S. and Ph.D degree in electrical and computer engineering from the University of California at Santa Barbara, in 1985 and 1990, respectively. He is currently the Yeager Families Professor with the College of Engineering, University of California at Riverside, USA. He is also serving as the Director for the Center for Environmental Research and Technology. His current research interests include ITS and the environment, transportation/emissions modeling, vehicle activity analysis, advanced navigation techniques, electric vehicle technology, and advanced sensing and control. Dr. Barth has been active in the IEEE Intelligent Transportation System Society for many years, serving as a Senior Editor for both the Transactions of ITS and the Transactions on Intelligent Vehicles. He served as the IEEE ITSS President for 2014 and 2015 and is currently the IEEE ITSS Vice President of Education.
\end{IEEEbiography}

\begin{IEEEbiography}
    [{\includegraphics[width=1in,height=1.25in,clip,keepaspectratio]{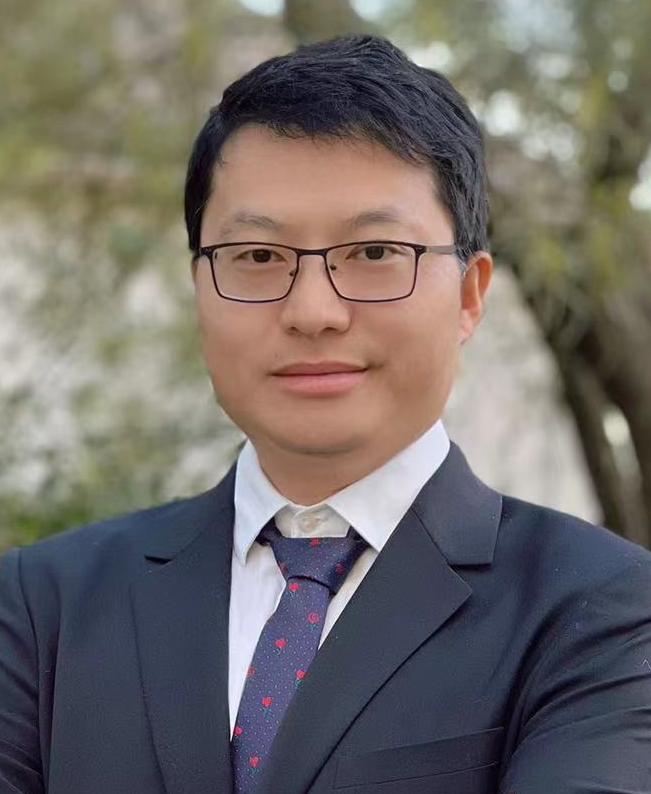}}]
    {Xuewei Qi}
(Member, IEEE) received his Ph.D. degree in electrical and computer engineering from the University of California-Riverside in 2016 and his M.S. degree in engineering from the University of Georgia, USA, in 2013. He is a Principle AI Researcher with Toyota North America Research Labs (Silicon Valley). He was with General Motors as an Artificial Intelligence and Machine Learning Research Scientist. He was also working as a Lead Perception Research Engineer at Aeye.ai. His recent research interests include deep learning, autonomous vehicles, perception and sensor fusion, reinforcement learning and decision making. He is also serving as a member of several standing committees of the Transportation Research Board (TRB).
\end{IEEEbiography}

\begin{IEEEbiography}
    [{\includegraphics[width=1in,height=1.25in,clip,keepaspectratio]{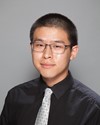}}]{Yongkang Liu}
received the Ph.D. and M.S. degrees in electrical engineering from the University of Texas at Dallas in 2021 and 2017, respectively. He is currently a Research Engineer at Toyota Motor North America, InfoTech Labs. His current research interests are focused on in-vehicle systems and advancements in inteligent vehicle technologies.  
\end{IEEEbiography}

\begin{IEEEbiography}
    [{\includegraphics[width=1in,height=1.25in,clip,keepaspectratio]{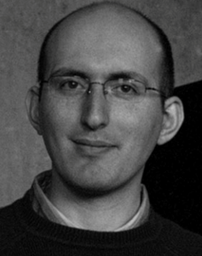}}]{Emrah Akin Sisbot}
(Member, IEEE) received the Ph.D. degree in robotics and artificial intelligence from Paul Sabatier University, Toulouse, France in 2008. He was a Postdoctoral Research Fellow at LAAS-CNRS, Toulouse, France, and at the University of Washington, Seattle. He is currently a Principal Engineer with Toyota Motor North America, InfoTech Labs, Mountain View, CA. His current research interests include real-time intelligent systems, robotics, and human-machine interaction.
\end{IEEEbiography}

\begin{IEEEbiography}
    [{\includegraphics[width=1in,height=1.25in,clip,keepaspectratio]{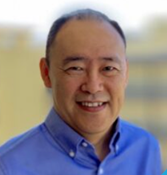}}]
{Kentaro Oguchi} received the M.S. degree in computer science from Nagoya University. He is currently a Director at Toyota Motor North America, InfoTech Labs. Oguchi’s team is responsible for creating intelligent connected vehicle architecture that takes advantage of novel AI technologies to provide real-time services to connected vehicles for smoother and efficient traffic, intelligent dynamic parking navigation and vehicle guidance to avoid risks from anomalous drivers. His team also creates technologies to form a vehicular cloud using Vehicle-to-Everything technologies. Prior, he worked as a senior researcher at Toyota Central R\&D Labs in Japan.
\end{IEEEbiography}




\end{document}